\documentclass[preprint,12pt,authoryear]{elsarticle}
\usepackage{soul,color}
\sethlcolor{yellow}
\usepackage{amssymb}
\usepackage{lineno}
\modulolinenumbers[5]

\usepackage{float}
\usepackage{graphicx}
\usepackage{amsmath}
\usepackage{subfigure}
\usepackage{hyperref}

\usepackage[ruled,linesnumbered]{algorithm2e}

\begin{document}

\begin{frontmatter}


\title{Representation Learning for Continuous Action Spaces is Beneficial for Efficient Policy Learning\tnoteref{mytitlenote1}}
\tnotetext[mytitlenote1]{This manuscript has not been published, was not, and is not being submitted to any other journal and conference.}

\author[1]{Tingting Zhao}
%\ead{tingting@tust.edu.cn}

\author[1]{Ying Wang}
%\ead{ilse_wang@163.com}

\author[1]{Wei Sun}
%\ead{sunweitust@mail.tust.edu.cn}

\author[1]{Yarui Chen\corref{cor1}}
\cortext[cor1]{Corresponding author}
%\ead{yrchen@tust.edu.cn}

\author[2]{Gang Niu}
%\ead{gang.niu.ml@gmail.com}

\author[2,3]{Masashi Sugiyama}
%\ead{sugi@k.u-tokyo.ac.jp}

\address[1]{College of Artificial Intelligence, Tianjin University of Science and Technology, Tianjin 300457, P.R.china}
\address[2]{RIKEN Center for Advanced Intelligence Project (AIP), Tokyo, Japan}
\address[3]{Graduate School of Frontier Sciences, The University of Tokyo, Tokyo, Japan}

\begin{abstract}
Deep reinforcement learning (DRL) breaks through the bottlenecks of traditional reinforcement learning (RL) with the help of the perception capability of deep learning and has been widely applied in real-world problems. While model-free RL, as a class of efficient DRL methods, performs the learning of state representations simultaneously with policy learning in an end-to-end manner when facing large-scale continuous state and action spaces.
However, training such a large policy model requires a large number of trajectory samples and training time. On the other hand, the learned policy often fails to generalize to large-scale action spaces, especially for the continuous action spaces.
To address this issue, in this paper we propose an efficient policy learning method in latent state and action spaces. More specifically, we extend the idea of state representations to action representations for better policy generalization capability. Meanwhile, we divide the whole learning task into learning with the large-scale representation models in an unsupervised manner and learning with the small-scale policy model in the RL manner. The small policy model facilitates policy learning, while not sacrificing generalization and expressiveness via the large representation model. Finally, the effectiveness of the proposed method is demonstrated by MountainCar, CarRacing and Cheetah experiments.
\end{abstract}

%%\includegraphics[scale=.42]{graphical abstracts}

%%\item An efficient policy learning framework in latent space (PL-LS) is proposed.
%%\item The action representation is learned in the policy model.
%%Idea of state representation in latent space is extended to action representation.
%%\item Building a compact policy model in PL-LS can reduce the burden of policy learning.
%%\item PL-LS can improve the  learning efficiency and generalization capability.
%%\item The performance is significantly improved.


\begin{keyword}
Policy model \sep Model-free reinforcement learning \sep Continuous action spaces \sep State representations \sep Action representations



\end{keyword}

\end{frontmatter}

%\linenumbers

\section{Introduction}

\emph{Deep reinforcement learning} (DRL) controls an agent's behavior directly through the learning of high-dimensional perceptual inputs by combining the perceptual capabilities of deep learning (DL) with the decision making capabilities of reinforcement learning (RL). In DRL, the goal of the agent is to learn a policy to maximize its cumulative rewards through interacting with the environment. So far, DRL has been successfully applied to many real-world tasks: e.g., robotics \citep{ICML:Peters+Schaal:2006,2017robotic}, games \citep{2013PlayingAtari,2016Mastering,2017Deep}, video prediction \citep{2015Action}, autonomous driving \citep{2012helicopters,2019Model}, intelligent transportation \citep{2014Intelligent,2021Hierarchical}, etc.

Benefitting from the large-scale policy model and excellent state perception capability, DRL breaks through the bottlenecks of traditional RL and could be applied to the real-world problems. Generally, policy learning in DRL could be categorized into two types. One is a value-function-based policy learning approach, which first learns the value function of the state-action pairs, and the action is chosen based on the estimated value function. The classical methods among this category include Deep Q-network (DQN) \citep{2019DQN}, Double DQN \citep{2016DoubleDQN}, Prioritized Experience Replay \citep{2015Prioritized}, and Dueling Architecture \citep{2016Dueling}. However, it is difficult to select actions by maximizing the value function at each time step due to extreme non-convexity of value function. Therefore, the value-function-based approach is more suitable for decision making tasks in discrete action spaces. The other type of policy learning approach is the policy-based algorithm, which directly models the policy and is suitable for solving complex decision tasks with continuous action spaces, such as policy gradient algorithm \citep{mach:Williams:1992}. Furthermore, the Actor-Critic (AC) architecture \citep{book:Sutton+Barto} combines the value-function-based approach and the policy-based approach, in which the actor plays the role of policy-based approach and is used to control how the agent behaves, while the critic evaluates  the current actor in terms of the value function approximation and guides the actor in policy improvement. The representative algorithms among AC framework include Deep Deterministic Policy Gradient (DDPG) \citep{2015Continuous}, Asynchronous Advantage Actor-Critic (A3C) \citep{2016A3C}, Soft Actor-Critic (SAC) \citep{SAC} and Proximal Policy Optimization (PPO) \citep{PPO2017}.

The key to the success of the aforementioned DRL methods is the ability to perceive high-dimensional states and the ability to represent complex tasks in deep decision-making models, as concluded in \cite{NIPS2017_3323fe11}. Extensive researches demonstrated that utilizing a range of deep neural network (DNN) architectures allows the agent to successfully learn control policies directly from high-dimensional sensory input using RL methods in the end-to-end manner \citep{2017Rainbow,2020endtoend,mayo2021visual}. However, a huge number of samples and training time are absolutely required to learn millions of weights of a large model in DRL algorithms. For example, the Rainbow model usesd 180 million frames, or 83 hours of game data to achieve 40 games over human performance on 57 Atari games \citep{2017Rainbow}; the model of AlphaStar AI underwent 44 days of training and beat 99.8\% of European players in the StarCraft 2 game \citep{2020Grandmaster}; OpenAI Five went through 10 months of real-world training to beat the Dota 2 human player's world champion (Team OG) \citep{2019Dota}. In addition, training DNNs by the end-to-end manner suffers from the problem of overfitting, and the learned model often fails to generalize to seemingly small changes in the environment \citep{NIPS2017_3323fe11}, which also increases the burden and difficulty of policy learning in RL.

In order to perform policy learning efficiently while satisfying the requirement for generalization ability, DRL algorithms usually leverage state representations for generalization \citep{PPO2017,2015Continuous,2015Embed}. Recent work has shown the benefits associated with state representations for large-scale and continuous state spaces \citep{worldmodels,2019Dream}, but relatively little research has been done on action representations for large-scale and continuous action spaces. In order to further improve the generalization capability over actions, we would like to extend the idea of learning latent representations of high-dimensional states to explore latent action representations in continuous action spaces, and investigate whether action representations also preserve the advantages as state representations. Therefore, the policy in the proposed latent action space will learn action representation rather than the raw action. Ideally, the proposed policy model can be generalized to other actions with similar representations, where similar values of  action representations are regarded as similar representations. Thus, the generalizaition capability over the action space is improved.

On the other hand, state representations with the proposed action representations may enable the agent to better understand the semantic of the state and action, and allow the agent to reason and learn over the learned latent spaces.
However, the representations of high-dimensional states in the traditional DRL algorithms are usually combined with policy learning together using an end-to-end learning manner, which makes the models for policy learning very large and the efficiency is not satisfied.
Therefore, we propose an efficient policy learning method in latent state and action spaces to reduce the burden of policy learning and improve the learning efficiency by constructing a small-scale and compact policy model with the help of the state and action representations in latent space, which is referred to as policy learning in latent spaces (PL-LS). More specifically, we abandon the end-to-end learning manner used in the traditional DRL algorithms and divide the whole learning task into the large representation models and the small policy model. We first learn  the state and action representation models in latent spaces in an unsupervised learning manner, and then learn the small-scale policy model  in the latent spaces for the RL problem. We believe this separation is desirable because unsupervised learning methods tend to be better performed and more reliable than RL methods \citep{worldmodels}.

To evaluate the effectiveness of our proposed framework, we conducted experiments on the MoutainCar, CarRacing and Cheetah tasks, showing that it is feasible to introduce action representations in latent space that not only
generalize the actions to action choices with similar representations, but also further explore the action and state spaces.
In addition, the policy learned based on our proposed framework is significantly improved with the help of state and action representations in latent space in terms of computational efficiency and performance.

The rest of this paper starts with a review of related work, and then introduces the formalization of RL and the generative model will be used to learn the state representations and action representations, i.e., the variational autoencoder (VAE) \citep{2016VAE}. Afterwards we explain the structure of our proposed model and the algorithmic procedure in detail. The experimental results are then presented and analyzed. Finally, we give the conclusion.

\section{Related Work}

In this section, we review the most related work, and discuss how they relate to our proposed method.

\paragraph*{State Representations} In DRL tasks, the powerful description and abstract representation of high-dimensional data by deep learning are usually applied to assist the automatic representation of states in DRL, which plays a strong supporting role in policy learning. We discuss and study the work related to state representations from two aspects. From the perspective of model-free DRL, state perceptions are combined with policy learning together to successfully learn the control policies from the raw observed states in an end-to-end learning manner, such as Deep Q-network \citep{2019DQN}, TRPO \citep{2015Trust}, and PPO \citep{PPO2017}. Therefore, a large-scale DNN needs to be built to well realize the perception and abstraction of states, but this will increase the demand for sample size and the difficulty of policy learning.

On the other hand, from the perspective of model-based DRL, in order to solve the problem of high-dimensional observed states, E2C \citep{2015Embed} and RCE \citep{2017Robust} convert the optimal control problem of high-dimensional nonlinear systems into a local linear problem in low-dimensional latent space. World Models \citep{worldmodels} use a visual sensory component to learn the transition from observed input frames to state representations, and abandon the end-to-end learning manner. Our approach is similar to it, but we introduce the non-end-to-end learning manner to model-free RL. We also use VAE \citep{2014Autobayes,2016VAE} combined with a convolutional neural network (CNN) \citep{2017Convolutional}, but we design a set of network models that match the data properties of RL itself to learn state representations in latent space. PlaNet \citep{2018Learning} is a fully model-based agent that learns environmental dynamics from images and selects actions by performing fast online planning in latent space. Dreamer \citep{2019Dream} and DreamerV2 \citep{2020Mastering} learn behavior purely from predictions in the compact latent space of the world model. Our approach has similarities to the Dreamer methods in that we also learn policies and predictions of value functions in latent space of learned states. However, we are not fully immersed in the imaginary space, since we focus on the model-free RL setup, where the agent is still required to interact with the real environment to ensure the accuracy of policy learning.

The employment of the reconstruction in the above methods for learning the latent state representations can greatly improve the data efficiency of RL. However, when encoding the high-dimensional observed states, the information that are not relevant to the task is also taken into account, resulting in redundant information being encoded and useless latent state representations being generated, thus affecting the attention of the agent. DeepMDP \citep{gelada2019deepmdp} introduces the idea of mutual simulation and reward prediction as well as latent state transition distribution prediction as auxiliary tasks to ensure that two states that are not relevant will not be encoded into the same state representation. However, DeepMDP relies on a strong assumption that the learned MDP representations are Lipschitz \citep{hinderer2005lipschitz,asadi2018lipschitz}. In contrast, Deep Bisimulation for Control (DBC) \citep{DBC} is guaranteed to generate the representation of the Lipschitz MDP by directly learning a state representation based on mutual simulation equivalence. \cite{DBC} argue that if one wants to learn a state representation that encodes only task-relevant information in the state and keeps task-irrelevant information constant, it is intuitively possible to determine task relevance through reward signals. Our approach provides a learning framework in which we can employ not only the traditional reconstruction based approach but also the mutual simulation equivalence based approach for the state representation.

\paragraph*{Action Representations} The existing DRL methods allow agents to use state representations for reasoning and learning. This idea was introduced into action representations, that is, learning a policy in latent space of action representations, and then mapping the action representations to the real action space, with less research related to them. Dulac-Arnold et al. proposed to embed discrete actions into a continuous space and then use the nearest neighbor method to find the optimal action \citep{2015Deep}. However, they assumed that the representation of the action is given as a priori, and do not provide a method for learning the action representations. Most relevant work would be the method of policy gradients with representations for actions (PG-RA)\citep{2019Learning}, which proposed to learn action representations as part of a policy structure and use a supervised way for its learning. However, their proposed method is restricted to tasks with discrete action spaces, and cannot be applied to deal with the continuous action problems naturally. It is the first attempt to learn action representations in latent space to deal with large-scale continuous action space problems, where the action representation is seperately learned from the policy learning. The correlations and similarities between actions are captured by action representations to enhance the generalization of action selections.
%In our proposed approach, on the other hand, the learning of action representations in latent space is partially independent of the policy model and can be used to deal with tasks in large-scale continuous action spaces. The correlations and similarities between actions are captured by action representations to enhance the generalization of action selections.%

\section{Background}

In this section, we start by introducing the notation and formalization of standard RL framework. Then, we describe the data representation model of variational autoencoder (VAE) that will be employed in the our proposed framework.

\subsection{Formulation of Reinforcement Learning}
In DRL, an agent interacts with an unknown environment to obtain an optimal policy so as to maximize the return, i.e., the cumulative discounted rewards \citep{book:Sutton+Barto}. Generally, we model interactions with the environment in a RL task as a \emph{Markov decision process} (MDP), which is described by the tuple of $(S, A, P_T, P(s_1), r)$, where
\begin{itemize}
	\item $S$ is the state space, which could be continuous or discrete, but we focus on continuous state space in this paper. $s_t \in S$ denotes the state of the agent at the time step $t$;
	\item $A$ is the action space, which could be continuous or discrete, but we focus on continuous action space in this paper. $a_t \in A$ denotes the action taken by the agent at the time step $t$;
	\item $P_T : S\times A \times S \rightarrow [0,1]$ is the transition probability density function. $p(s_{t+1}|s_t,a_t)$ is the conditional probability density that the agent in state $s_t$ performs an action $a_t$ and transfers to the next state $s_{t+1}$;
	\item $P(s_1)$ denotes the probability density function of the initial state $s_1$;
	\item $r(s_t,a_t,s_{t+1})$ denotes the immediate reward for transfering to the next state $s_{t + 1}$ after performing the action $a_t$ at the current state $s_t$;
\end{itemize}

The agent starts from the initial state $s_1$ drawn according to $P(s_1)$. At each time step $t$, the agent senses its current state $s_t$ and selects an action $a_t$ following a policy $\pi(a_t|s_t)$, which yields the agent’s behavior, i.e., a mapping from state $s_t$ to action $a_t$, then transfers to the next state $s_{t+1}$ according to the state transition probability density $p(s_{t+1}|s_t,a_t)$, and receives a reward $r(s_t,a_t,s_{t+1})$ for this state transition. The agent repeats the above process until it reaches a terminal state or the maximum time step $T$ , which is referred to as a trajectory. The probability density of the occurred trajectory is \[p(h):=p(s_1)\prod_{t=1}^{T}p(s_{t+1}|s_t,a_t)\pi(a_t|s_t).\]
Once a trajectory is obtained, the return of this trajectory can be calculated: \[R(h):=\sum_{t=1}^{T}\gamma^{t-1}r(s_t,a_t,s_{t+1}),\] where $\gamma \in (0,1]$ is the reward discounting parameter that is used to weight the impact of future rewards. We measure the quality of a policy by the expectation of the return, which can be expressed as \[J_{\pi}:=\int p(h)R(h)dh.\] The goal of RL is to find an optimal policy $\pi^*$ that maximizes the expected return $J_{\pi}$: \[\pi^*:=\arg\max_{\pi} J_{\pi}.\]

\subsection{Variational Autoencoder}
Representation learning learns the implicit features of data automatically \citep{2012Representation}, and is typically used to extract useful information when building classifiers or other predictors. Representation learning has been widely applied in tasks such as speech recognition \citep{2011Context}, target recognition \citep{2012ImageNet}, and natural language processing \citep{2012Joint}. The autoencoder is one of the classic representation learning methods, which processes complex high-dimensional data in an unsupervised way \citep{1986autoencoder} and realizes the data representation and reconstruction by its encoder and decoder. The encoder learns the latent characteristics of input data and decoder reconstructs the abstracted data. The autoencoder is often used as a method of feature extraction, and has been widely used in image classification \citep{2017Clustering}, video anomaly detection \citep{2016Video, 2020Clustering}, and pattern recognition \citep{2014Feature}.

The variational autoencoder (VAE) combines deep learning with Bayesian inference while maintaining the basic functionalities of the autoencoder \citep{2014Autobayes,2016VAE}. The traditional autoencoder captures the latent features of the data in a numerical form by the encoder, while VAE obtains the probability distribution of the latent features of the data, which greatly improves the generalization ability of generated data.

\begin{figure}[H]
	\centering
	\includegraphics[scale=.52]{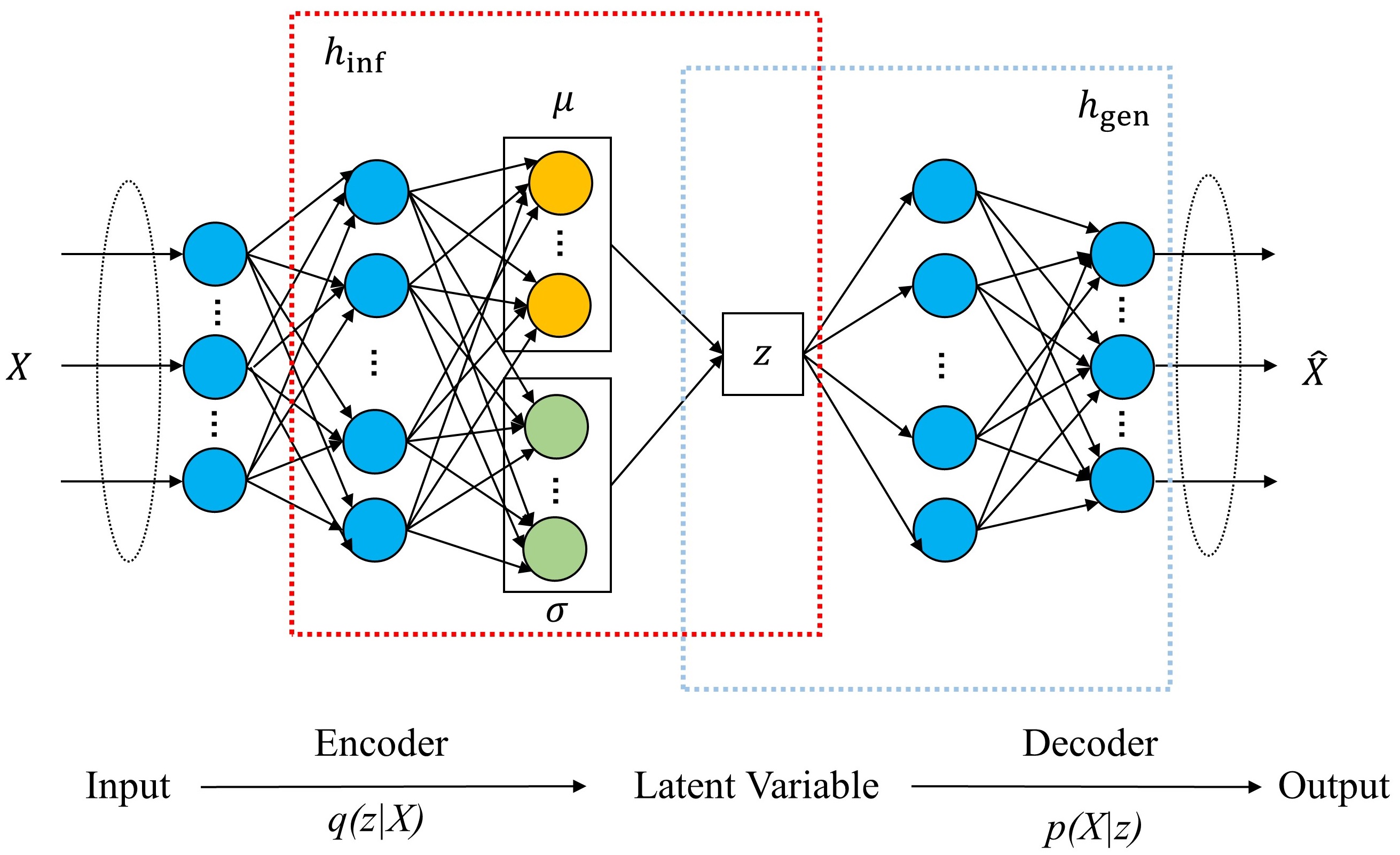}
	\caption{The overall structure of VAE. VAE consists of an inference network $h_{\mathrm{inf}}$ and a generation network $h_{\mathrm{gen}}$, which completes the process of encoding and decoding the input data $X$ to obtain the reconstructed data $\widehat{X}$.} \label{vae_structure}
\end{figure}

The structure of VAE is shown in \hyperref[vae_structure]{Fig.1}, where $X$ is the input data, $h_{\mathrm{inf}}$ is the inference network used to obtain the data distribution in latent space, $\mu$ and $\sigma$ respectively represent the mean and standard deviation of the data distribution in latent space, $z$ is the latent representation of the original data in latent space, namely the latent variable; $h_{\mathrm{gen}}$ is the generation neural network used to generate new data, $\widehat{X}$ is the new data generated by the generation network; $q(z|X)$ and $p(X|z)$ refer to the learned conditional densities of the latent variable and the original data during the encoding and decoding process, respectively.
In order to make VAE with data generation capability rather than deterministic mapping relationship, the latent variable $z$ is required to be a random variable, which is usually assumed to follow the multivariate normal distribution, i.e., $p(z) \sim N(0,I)$, where $I$ is the identity matrix.

The objective of VAE is to maximize the likelihood of the generated data while simultaneously minimizing the distance between the prior distribution of the latent variable and the inference model, which is formulated as follows: \begin{equation} \label{Eq.1}
	L_{\mathrm{VAE}} = \mathrm{KL}(q(z|X)||p(z))-E_{z\sim q(z|X)}[\log p(X|z)],
\end{equation} where KL refers to the Kullback-Leibler divergence of $p$ from $q$;
the second term in \hyperref[Eq.1]{Eq.~(1)} measures the reconstruction error, and the first term in \hyperref[Eq.1]{Eq.~(1)} is the additional constraint on the latent representations. VAE optimizes the loss function so that the estimated density $q(z|X)$ is close to $p(z)$, and the reconstruction error is expected to be small.

In order to avoid the no-gradient problem caused by random sampling of the latent variable $z$, VAE uses the reparameterization technique to introduce the parameter $\varepsilon \sim N(0,I)$, which transforms the direct sampling of the latent variable $z$ into a linear operation of $z=\mu+\sigma*\varepsilon$, and enables it to be optimized using the gradient descent algorithm.

During the training procedure, the inference model $q(z|X)$ firstly extracts the mean and standard deviation of the distribution of the latent variable $z$ corresponding to the sample $X$. Then, the reparameterization technique is used to obtain the latent variable $z$. Finally, the generation model $p(X|z)$ outputs the reconstructed sample $\widehat{X}$ corresponding the given latent variable $z$.

\section{Efficient Policy Learning in Latent State and Action Spaces}
\begin{figure}[H]
	\centering
	\includegraphics[scale=.65]{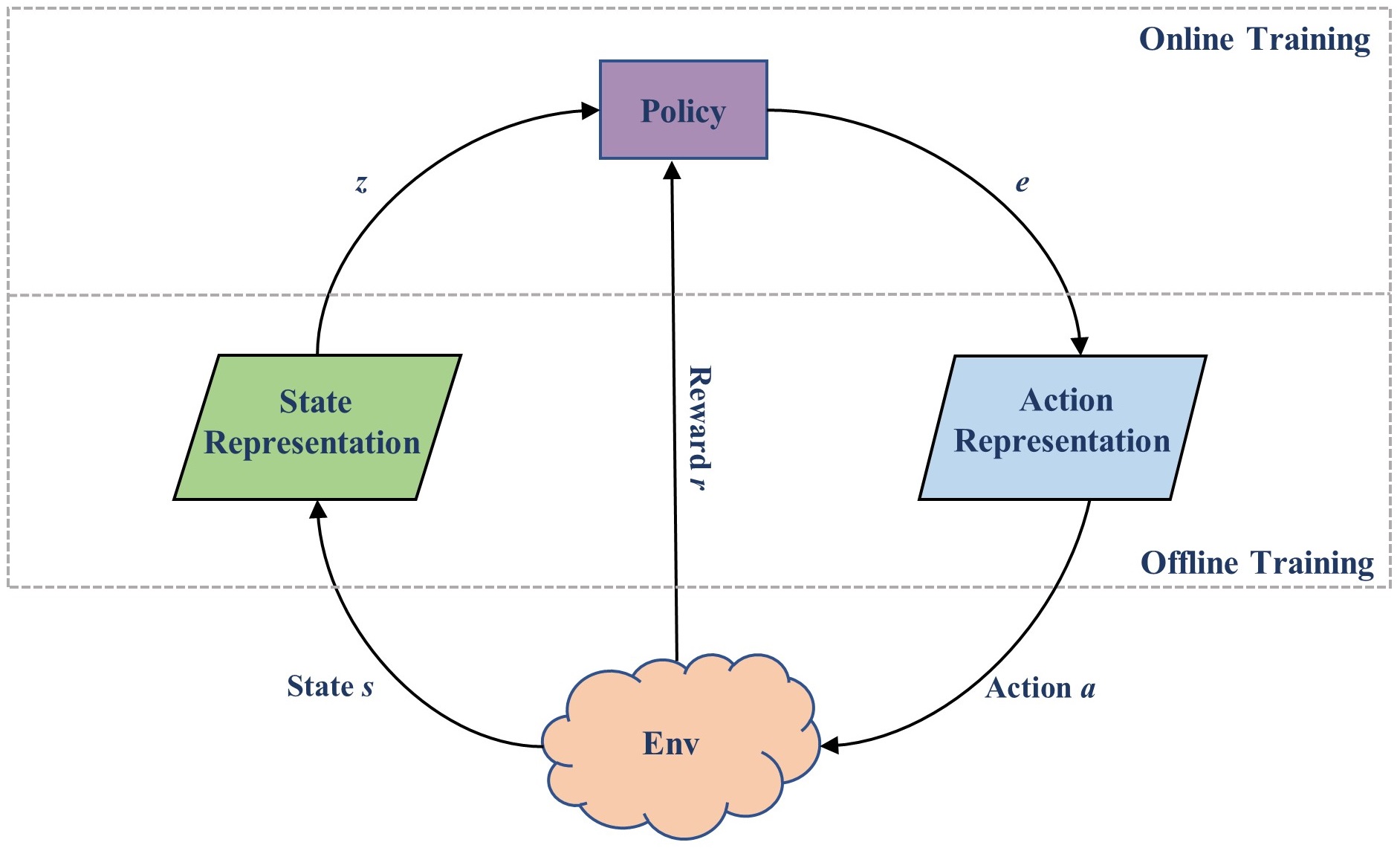}
	\caption{The overall framework of the proposed efficient policy learning model. The state and action representations in latent spaces and the policy learning make up the whole model, where the state and action representations are trained in the offline manner and the policy is learned in the online manner.} \label{overall_structure}
\end{figure}

Facing a complex environment, in order to solve a large-scale decision-making problem in continuous state and action spaces more efficiently, we try to alleviate the problems of low learning efficiency and weak generalization ability in the field of DRL from three aspects: state representation, action representation and compact policy learning. The overall structure of the proposed framework is shown in \hyperref[overall_structure]{Fig.2}, where we divide the learning process into the offline learning and online learning. More specifically, we first leave the large-scale network training to the offline procedure in an unsupervised learning way, where we learn the state representation and action representation in their latent spaces; then, we learn a small-scale and compact policy model online by the RL algorithm.

%In this structure, we use offline learning to learn the representation of state in low dimensional latent space and the prediction of state representation of future state in latent space; Using offline learning, we learn the mapping relationship between action representations in latent space and raw actions; Finally, a small number of real samples are used to learn an efficient and compact policy model online, so that it can quickly adapt to the current environment and tasks.

\subsection{State Representation in Latent Space}
\begin{figure}[H]
	\centering
	\includegraphics[scale=.60]{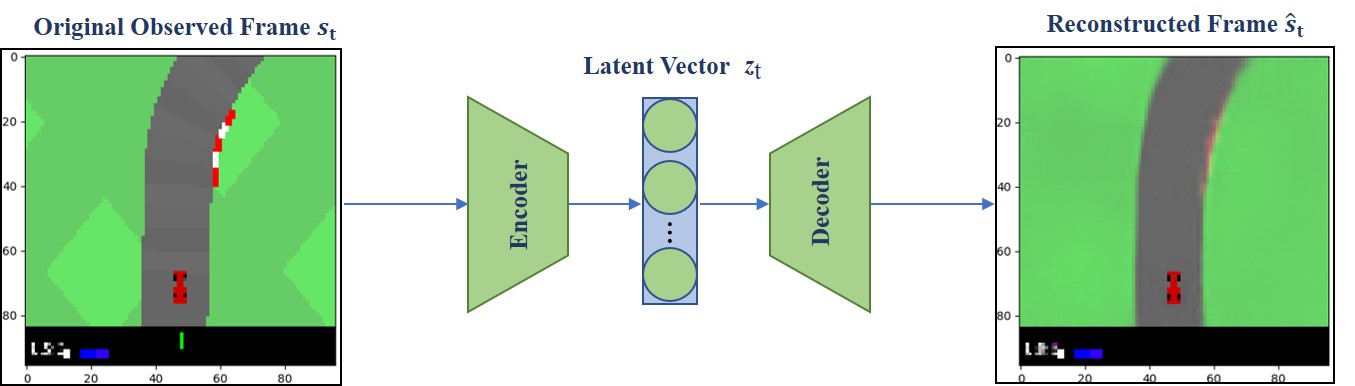}
	\caption{The framework of VAE-based state representation VAE($s_t$), abbreviated as $V_s$. At each time step, the original observed state $s_t$ is input into the encoder of $V_s$, and the low-dimensional state representation, i.e., latent vector $z_t$, is obtained. With the decoder of $V_s$, the state representation $z_t$ could be reconstructed to $\hat{s}_t$.} \label{state_representation}
\end{figure}

In order to effectively handle the high-dimensional state problems in large-scale environment and relieve the burden of subsequent policy learning, we intend to use the representation learning to abstract the high-dimensional observed state and make the agent to reason and learn using the represented state rather than the raw state. In this paper, we establish a large-scale network in line with the characteristics of state data to compress each observed state into a small latent vector $z$ as shown in \hyperref[state_representation]{Fig.3}, where we employ the well researched model of VAE. More specifically, the original observed state $s_t$ is input to the encoder, and then the latent vector $z_t$ is obtained, and finally $z_t$ is passed through the decoder to obtain the reconstructed frame $\hat{s_t}$.
This VAE-based state representation model is denoted by VAE($s_t$), and we refer to $V_s$ for brevity.
%and predict the representation in latent space corresponding to the state generated in the future environment, as shown in \hyperref[state_representation]{Fig.3}.
The loss function of the state representation based on VAE is formulated as
\begin{equation} \label{Eq.2}
	L(s_t) =\mathrm{KL}(q(z_t|s_t)||p(z_t))-E_{z_t\sim q(z_t|s_t)}[\log p(s_t|z_t)],
\end{equation} where $s_t$ is the state variable and $z_t$ is the latent variable with respect to the observed state; $q(z_t|s_t)$ and $p(s_t|z_t)$ correspond to the conditional probability distributions in the encoding and decoding processes, respectively. Through the off-line learning, we get the learned encoder of $V_s$, and thus obtain the low-dimensional state representation in latent space.

\subsection{Action Representation in Latent Space}
In order to efficiently carry out policy learning and meet the requirements of generalization ability, RL usually places its hopes on state representations.
Learning the underlying state representation is a well researched and widely employed idea in the field of DRL, where the state representation improves the generalization ability across the state space. We extend the idea of state representation to action representation with expectation that there exists the underlying latent representation of action space, which could further accelerate policy learning and improve the generalization across the action space.

Following this line, we propose to learn a policy in the latent space of action representation, and then map the action representation to the original action space. Specifically, we introduce an action representation space $\xi\subseteq\mathbb{R}^d$. The agent learns the policy in this latent space, that is, the policy defines the probability density function of the action representation given the current state representation: $\pi\left(\left.e_t\right|z_t\right)$, where $z_t$ and $e_t$ represent state $s_t$ and action $a_t$ in latent space respectively. Along with this policy model in the latent space, we also learn a mapping function from action representation to real action. Then, the generated action will be used to interact with the environment.
Action representations are expected to be generalized to other actions with similar representations, which greatly improves the generalization performance of action selections.

\begin{figure}[H]
	\centering
	\includegraphics[scale=.65]{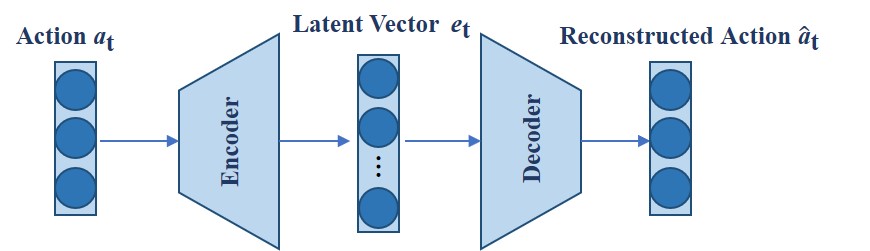}
	\caption{The framework of VAE-based action representation VAE($a_t$), abbreviated as $V_a$. At each time step, the original action $a_t$ is input into the encoder of $V_a$, and obtains the action representation, i.e., latent vector $e_t$. With the decoder of $V_a$, the action representation $e_t$ could be reconstructed to $\hat{a}_t$.} \label{actionModel}
\end{figure}

Similar to VAE-based state representation VAE($s_t$), we construct a VAE-based action representation model VAE($a_t$) shown in \hyperref[actionModel]{Fig.4}, abbreviated as $V_a$, where the decoder transforms action representations to real actions. The loss function of the action representation is described as follows:
\begin{equation} \label{Eq.3}
	L(a_t) =\mathrm{KL}(q(e_t|a_t)||p(e_t))-E_{e_t\sim q(e_t|a_t)}[\log p(a_t|e_t)],
\end{equation} where $a_t$ is the action variable and $e_t$ is the latent variable with respect to the action;  $q(e_t|a_t)$ and $p(a_t|e_t)$ correspond to the conditional probability distributions of action representation and reconstructed action in the encoding and decoding processes, respectively. We train the model  VAE($a_t$) in the offline manner. Finally, action representation in the latent space is obtained by the encoder of $V_a$, which will be used in the policy learning, and the transformation of the action representation to the actual action can be obtained by the decoder of $V_a$.

For the DRL tasks, the dimensionality of the state $s_t$ is usually high, while the dimensionality of the action $a_t$ is relatively low. For example, in the car-racing task, the observed state is the raw image with thousands of pixels, which is high-dimensional and information redundant; while the action is a 3-dimensional vector, consists of turning the steering wheel, stepping on the accelerator, and braking, where each dimension is linearly indistinguishable and informationally thin. Thus, we usually abstract the state and use a lower-dimensional latent state as its representation. On the other hand, we could also lift the action to a higher dimension in latent space to find its instructive representation \citep{zhang2021deformable}.

\begin{figure}[H]
	\centering
	\includegraphics[scale=.68]{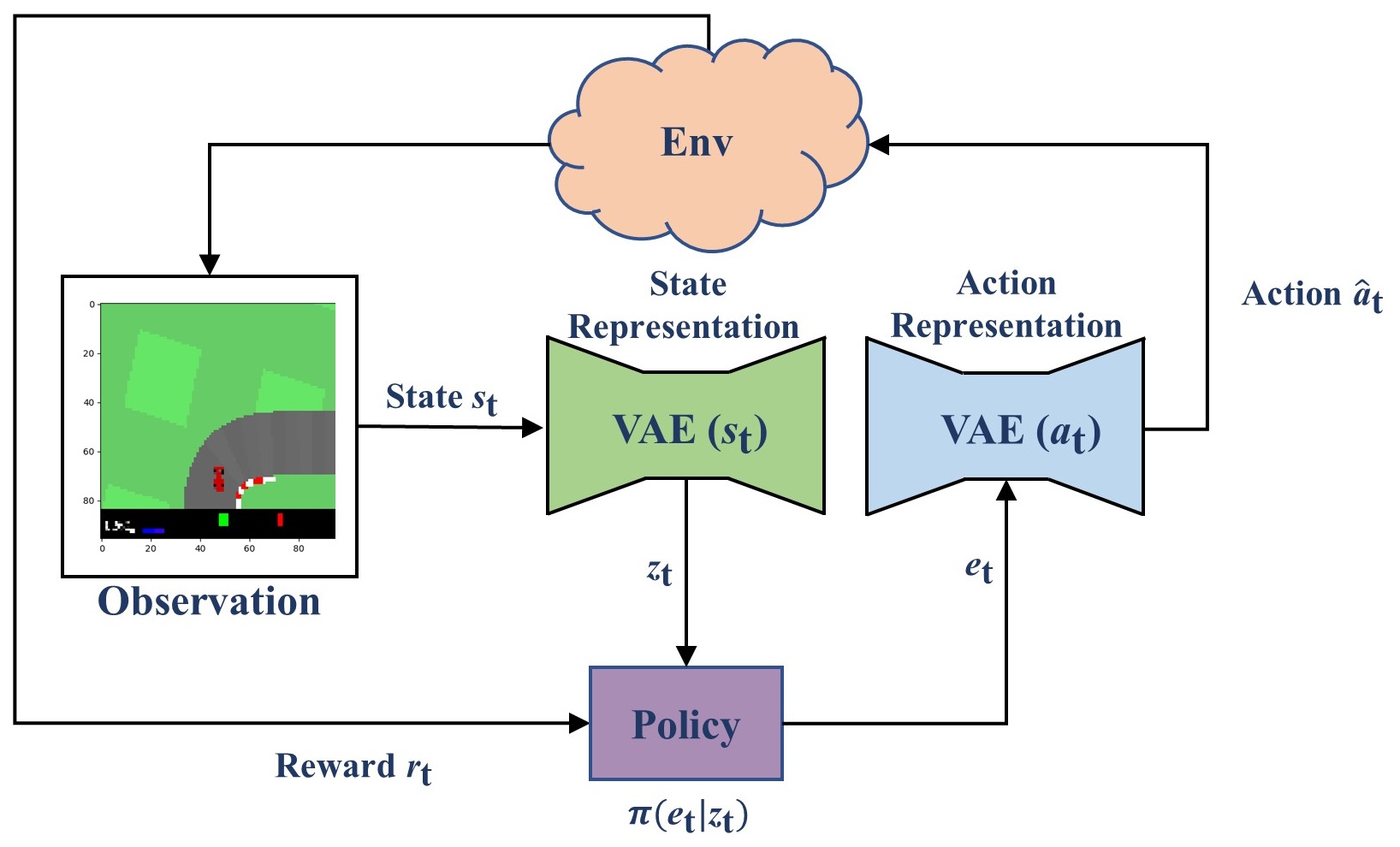}
	\caption{Flow chart of the proposed model of policy learning in latent spaces. At each time step, the raw observation $s_t$ is first processed by $V_s$ to obtain its state representation $z_t$, which is input to the new proposed policy model $\pi(e_t|z_t)$. Policy outputs an action representation $e_t$ in latent space, which will generate a corresponding action $\hat{a}_t$ through the decoder of $V_a$, and then affect the environment. Finally, the environment will produce an immediate reward $r_t$ and transit to the next state $s_ {t+1}$.} \label{all_flow_chart}
\end{figure}
\subsection{Algorithm of Policy Learning in Latent Spaces}
With the state representations and action representations, we could construct a small-scale policy model and learn it efficiently. We illustrate how state representations and action representations in latent spaces interact with the environment in the flow chart of \hyperref[all_flow_chart]{Fig.5}.
At each time step, the agent receives the state representation $z_t$ from the encoder of VAE($s_t$), which is input to the policy $\pi$ and gets the action representation $e_t$. The action representation $e_t$ goes through the decoder of VAE($a_t$), the reconstructed action $\hat{a}_t$ is obtained and used to interact with the environment. Subsequently, the agent transits to the next state $s_{t+1}$ and receives a reward $r(s_t,\hat{a}_t,s_{t+1})$. The above process is repeated for $T$ times, and a trajectory is obtained.

We implement the proposed latent space based policy model in the framework of policy-based algorithms to deal with the large-scale and continuous state and action problems. The algorithm is called the \textbf{P}olicy \textbf{L}earning algorithm in \textbf{L}atent \textbf{S}paces (PL-LS), which is explicitly described in \hyperref[Algorithm 1]{Algorithm 1}. PL-LS first collects transition samples ${\lbrace(s_t,a_t)\rbrace}^T_{t=1}$ using a random policy, and then learns $V_s$ and $V_a$ by \hyperref[Eq.2]{Eq.~(2)} and \hyperref[Eq.3]{Eq.~(3)} using the collected samples in order to get the state and action representations in latent spaces, respectively. Once the state and action representations in latent spaces are learned, we can construct a compact and small-scale policy model and learn it efficiently.

\begin{algorithm}[H]
	\caption{Policy Learning Algorithm in Latent Spaces (PL-LS)} \label{Algorithm 1}
	\LinesNumbered
	Collect transition samples $\lbrace(s_t, a_t)\rbrace^T_{t=1}$ using a random policy\;	
	Learn $V_s$ by \hyperref[Eq.2]{Eq.~(2)} using the collected data $\lbrace(s_t, a_t)\rbrace^T_{t=1}$ to get the state representations in latent space\;
	Learn $V_a$ by \hyperref[Eq.3]{Eq.~(3)} using the collected data $\lbrace(s_t, a_t)\rbrace^T_{t=1}$ to get the action representations in latent space\;
	Initialize learning parameters\;
	\For{episode = 0,1,2,…}{
		Sample initial state $s_1$ from $P(s_1)$\;
		\For{t=1,2,…}{
			$z_t$=$V_s$.encoder($s_t$)\;
			Sample action representation in latent space, $e_t$, from $\pi(\cdot|z_t)$\;
			$a_t$=$V_a$.decoder($e_t$)\;
			Execute $a_t$ and observe $s_{t+1}, r_t$\;
			Update $\pi$ using \emph{any} policy search algorithm\;
		}
	}
	
\end{algorithm}

In \hyperref[Algorithm 1]{Algorithm 1}, Lines 5-14 illustrate the online update procedure for all of the parameters RL involves. Each time step in the episode is represented by $t$. For each episode, we need to sample initial state $s_1$ from $P(s_1)$. For each step, a state is first encoded to a state representation by the encoder of $V_s$. An action representation is sampled and is then mapped to an action by the decoder of $V_a$. Having executed this action in the environment, the immediate reward is then used to update the policy, $\pi$, using any policy-based algorithm. Note that, the proposed method is a general framework of policy learning in latent spaces. The policy update algorithm could be chosen accordingly in Line 12 in \hyperref[Algorithm 1]{Algorithm 1},  where the value-function-based policy learning approach could also be applied according to the specific task.

\section{Experimental Results}

The essential motivation of this work is to provide a policy model based on the state and action representations, which can be combined with the existing DRL algorithms to improve the efficiency of policy learning and the ability of generalization. In this section, we conduct experiments on three tasks to verify the effectiveness of our proposed method: MountainCar to explore the underlying structure of action representation in the learned latent space and verify the feasibility of introducing action representation; Car Racing task to investigate the effectiveness and applicability of efficient policy learning based on state representation and action representation; Cheetah task to validate the extensibility of our proposed framework.

\subsection{MountainCar}
The MountainCar task consists of two hills and a car. The goal is to get the power to reach the target position on top of the right hill through policy learning (note that the car's engine is not strong enough to scale the hill in a single pass). The state space $S$ is two-dimensional and continuous, which is consist of the position $x \in [-1.2, 0.5]$ and the velocity $\mathop{x}\limits^{.} \in [-0.07, 0.07]$, i.e., $s=(x, \mathop{x}\limits^{.})$. The target position is set at 0.45. The action space $A$ is one-dimensional and continuous, $a\in[-1, 1]$, which corresponds to the force applied to the car. When $a>0$, it means that the right force is applied to the car; When $a<0$, it means that the left force is applied to the car. The reward function is defined as
\begin{equation}
	r\left(s_{t}, a_{t}, s_{t+1}\right)= \begin{cases}-\left(\mathrm{a}_{t}\right)^{2} * 0.1+100, & \text { if } x_{t+1} \geqslant 0.45 \\ -\left(\mathrm{a}_{t}\right)^{2} * 0.1, & \text { otherwise }\end{cases}
\end{equation}
The defined reward function increases the difficulty of policy learning, because if the car fails to reach the target as soon as possible, it will find that it is best not to move, then it will no longer reach the target position.

In this task, we mainly verify the validity of introduced action representation in latent space and explore the underlying relationships between action representations. We do not carry out the state representations in latent space mainly because the state space in this task is only two-dimensional and the dimension is very low. On the other hand, the validity of state representations has been well investigated \cite{worldmodels, 2015Embed}.

\begin{figure}[H]
	\centering
	\includegraphics[scale=.60]{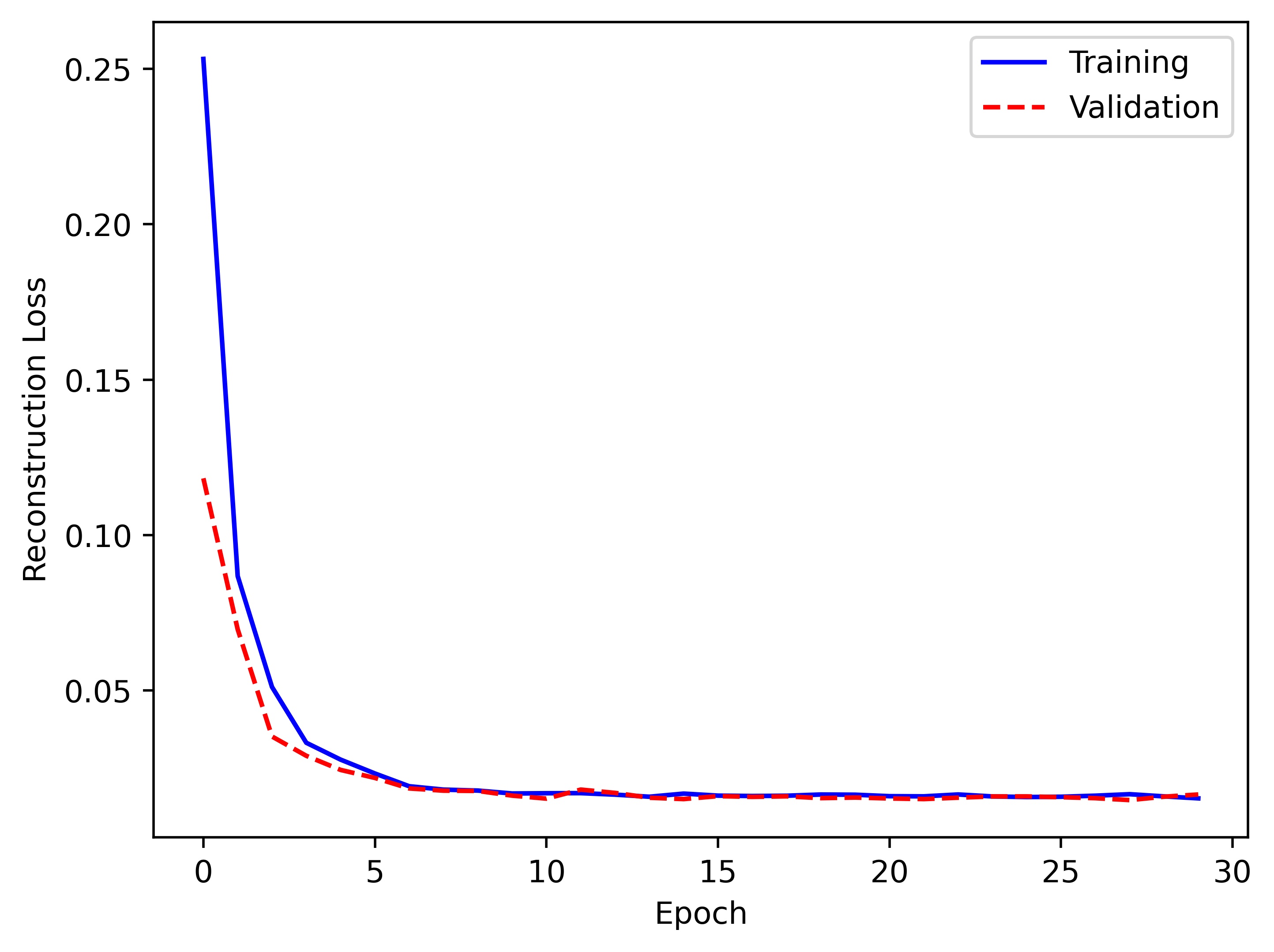}
	\caption{Learning process of the VAE-based action representation model $V_a$. The horizontal coordinate indicates the number of epochs during training, and the vertical coordinate indicates the reconstruction loss. The blue line indicates the reconstruction loss during training and the red line indicates the reconstruction loss during validation.} \label{VAE_loss}
\end{figure}

\paragraph*{Action Representation Learning}
In the MountainCar task, we apply the proposed VAE-based action representation VAE$(a_t)$, abbreviated as $V_a$, to capture the structure of the underlying action space. We use a random policy to make the car interact with the environment to collect data for 12 trajectories with the length of 999, of which 8400 action data are used as the training set and 3588 action data are used as the test set. Based on our preliminary experiments, we stretch one-dimensional action to three dimensional latent vector by the encoder of $V_a$, where the structure of encoder is designed as three-layer fully connected neural network, and the units in each layer are 32, 16 and 8 respectively. The latent vectors are then decoded and reconstructed through three-layer fully connected neural network with the units of 8, 16 and 32 in each layer respectively. We use mini-batch gradient descent to train the $V_a$ model with the Adam optimizer and the learning rate of 0.001. The learning process is shown in \hyperref[VAE_loss]{Fig.6}, where the model of $V_a$ gets converged round 10 epochs. We evaluate the performance of the learned $V_a$ model on the test data by the mean square error (MSE), which is $0.01553 \pm 0.00051$.

\paragraph*{Explore Action Representations in Latent Space}
In this section, we use the learned $V_a$ to explore the underlying structure of action representations. We set the initial state of the car as $s=(-1.2, 0)$, and sample 1000 actions uniformly in the action space $\boldmath{a}=[a_1, a_2,..., a_{1000}]$.

\begin{figure}[H]
	\centering
	\subfigure[Learned action representations with the partition of  500:500]{\includegraphics[width=0.47\columnwidth]{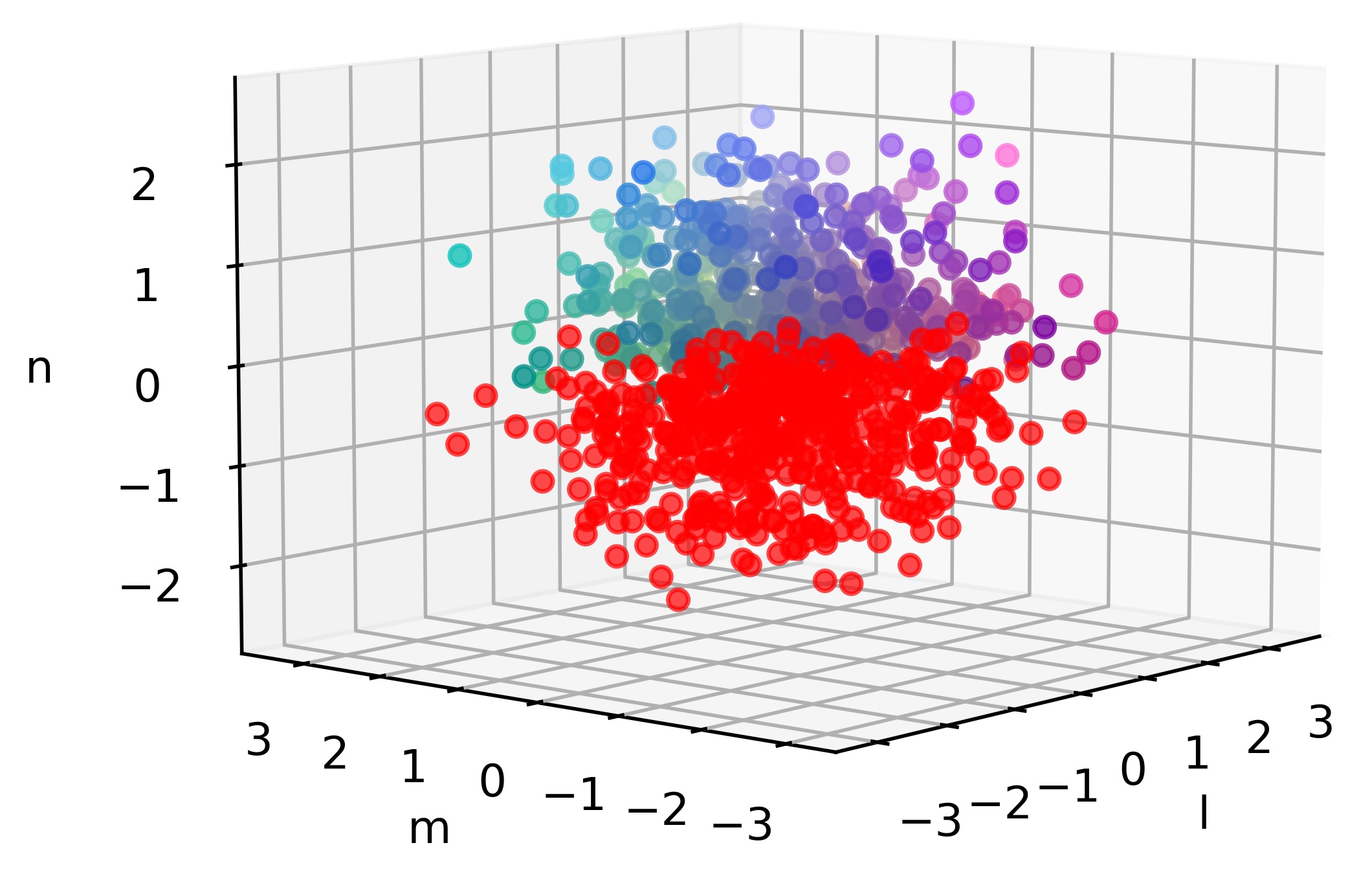}\label{MultiEmbedding500}}
	\subfigure[Learned action representations with the partition of  400:200:400]{\includegraphics[width=0.47\columnwidth]{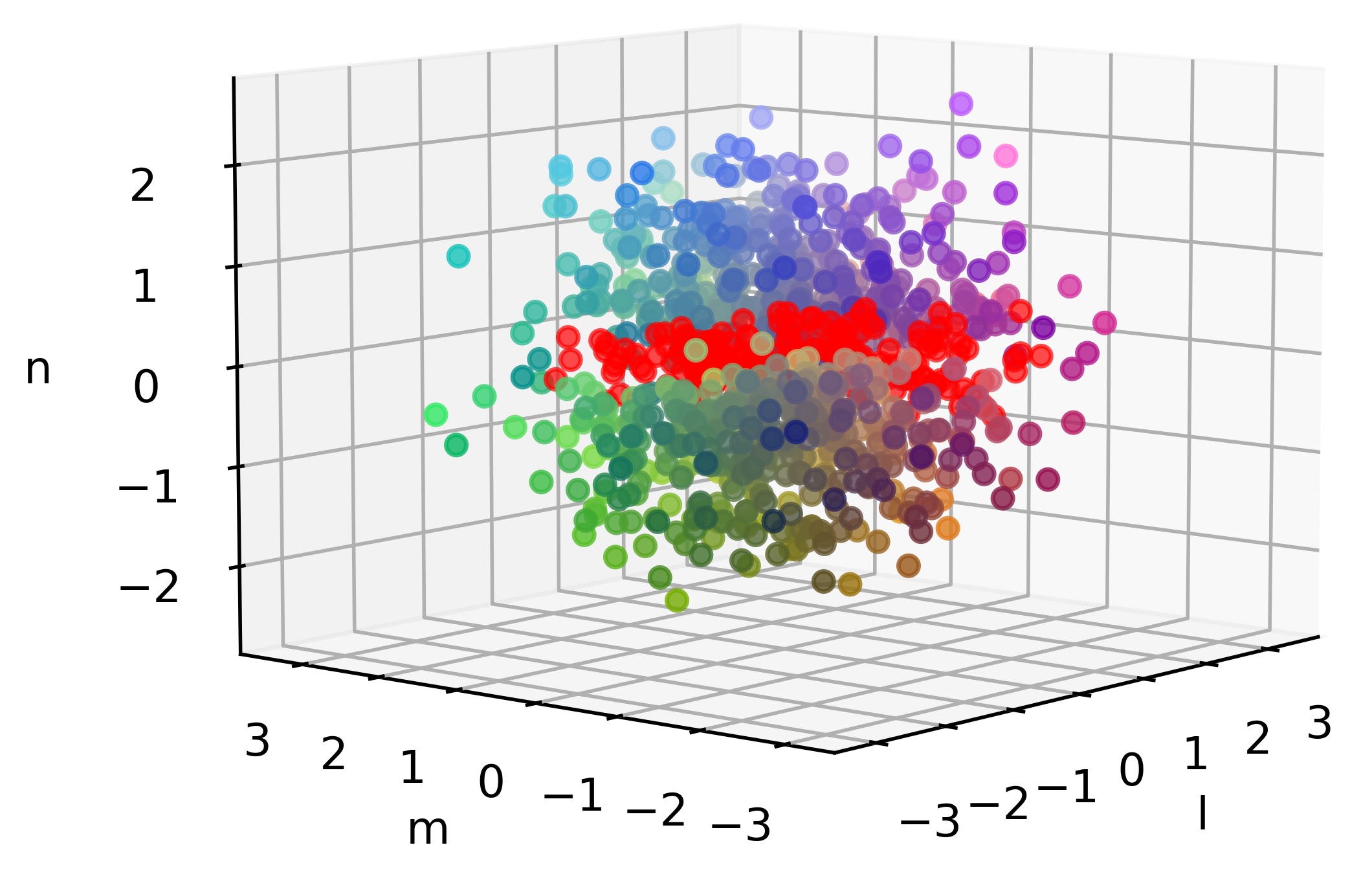}\label{MultiEmbedding200}}
	\subfigure[The effect of raw actions on the state]{\includegraphics[width=0.47\columnwidth]{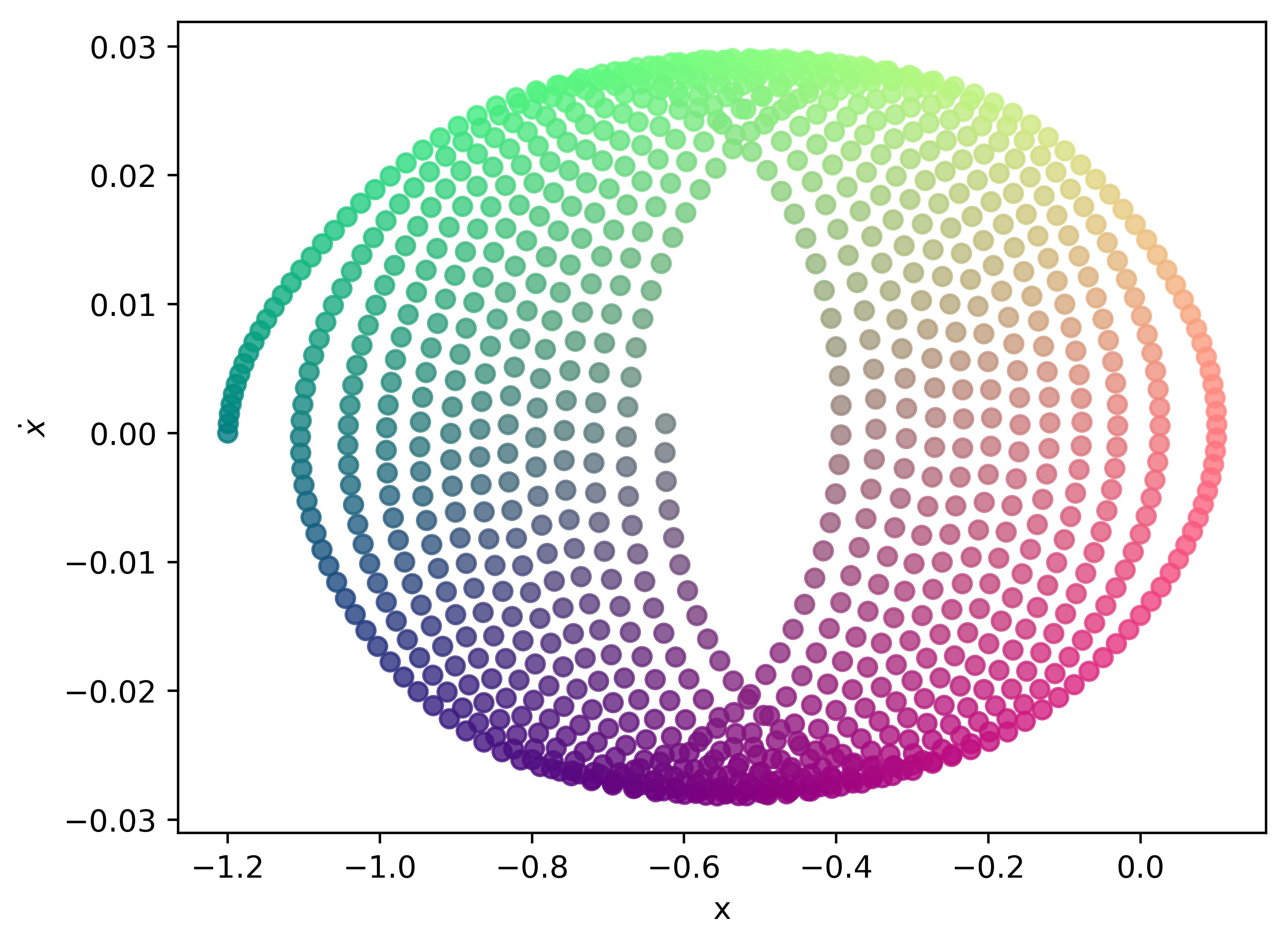}\label{MultiRawState}}
	\subfigure[The effect of reconstructed actions on the state]{\includegraphics[width=0.47\columnwidth]{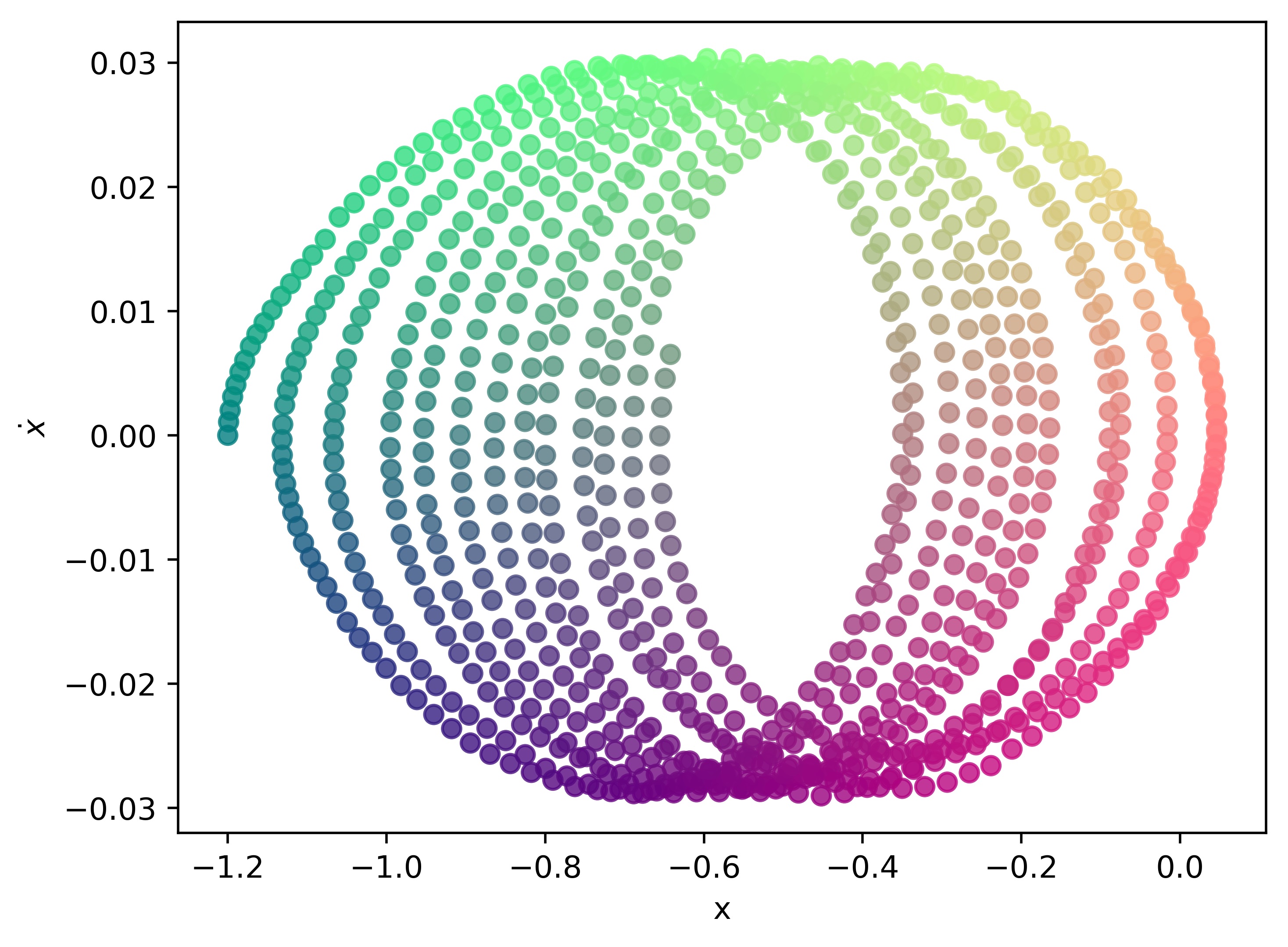}\label{MultiReconState}}
	
	\caption{Multi-step exploration of action representations in latent space. In Fig.7(a) and (b), the $l, m, n$ axes correspond to the three dimensions of action representations in the latent space, respectively. We take out the action representations in a specific area to mark it in red. Action representations in the other areas is colored based on the value $[l, m, n]$ of each dimension of the action representation, i.e., with the color $[R=l, G=m, B=n]$, where $l$, $m$ and $n$ are normalized to $[0,1]$.
		In Fig.7(c) and (d), the axes of $x$ and $\dot{x}$ correspond to the two dimensions of the state space, i.e., position and velocity, respectively. The effect of the reconstructed action and the effect of the original action on the state are illustrated, which demonstrate that the effect of the reconstructed actions on the state is similar to the original actions.}
	
	\label{Multi-step}
\end{figure}

Firstly, we investigate the underlying structure and the effect of action representations in the scenario of multi-step and continuous decision making. To achieve so, we exert those sampled actions to the car continuously, and obtains a trajectory $h$ with a length of 1000. Then, the sampled actions are encoded into the latent vectors $\boldmath{e}=[e_1, e_2,..., e_{1000}]$ through the encoder of the learned $V_a$. The latent vectors $e$ are decoded into the reconstructed action $\hat{\boldmath{a}}_t$ through the decoder of the learned $V_a$. Finally, the reconstructed actions could be used to interact with the environment.
The action representations of the sampled actions are illustrated  in \hyperref[MultiEmbedding500]{Fig.7(a)} and \hyperref[MultiEmbedding200]{Fig.7(b)}. In \hyperref[MultiEmbedding500]{Fig.7(a)}, we color the action representations $[e_{501}, e_{502},..., e_{1000}]$ corresponding to actions $[a_{501}, a_{502},..., a_{1000}]$ with red, the other points of action representations are colored according to the value of each dimension of the action representations $[l, m, n]$, i.e., the color is given by $[R=l, G=m, B=n]$, where $l$, $m$ and $n$ are the values of the elements in action representation and normalized to $[0, 1]$. In \hyperref[MultiEmbedding200]{Fig.7(b)}, we take out the action representations $[e_{401}, e_{402},...,e_{600}]$ corresponding to these actions $[a_{401}, a_{402},…, a_{600}]$, and color these action representations in red. Through the results in  \hyperref[MultiEmbedding500]{Fig.7(a)} and \hyperref[MultiEmbedding200]{Fig.7(b)}, the red points demonstrate that neighbouring points in the original action space also have similar structures in the latent space, that is, the corresponding action representations are also neighbouring. On the other hand, the other colored points show that color transition of the learned action representations in the latent space is smooth, which well demonstrates that the latent space preserves the relative underlying structure.

We further illustrate a comparison between the effect of the reconstructed actions on the state and the effect of the original actions on the state, in \hyperref[MultiRawState]{Fig.7(c)} and \hyperref[MultiReconState]{Fig.7(d)}. It can be seen that the effect of the reconstructed actions on the state is similar to the effect of the original actions on the state. The results show that the association relationship and underlying structure between the raw actions can be captured by using action representations, which shows that it is feasible to introduce the action represented latent space.

\begin{figure}[H]
	\centering
	\subfigure[Learned action representations]{\includegraphics[width=0.47\columnwidth]{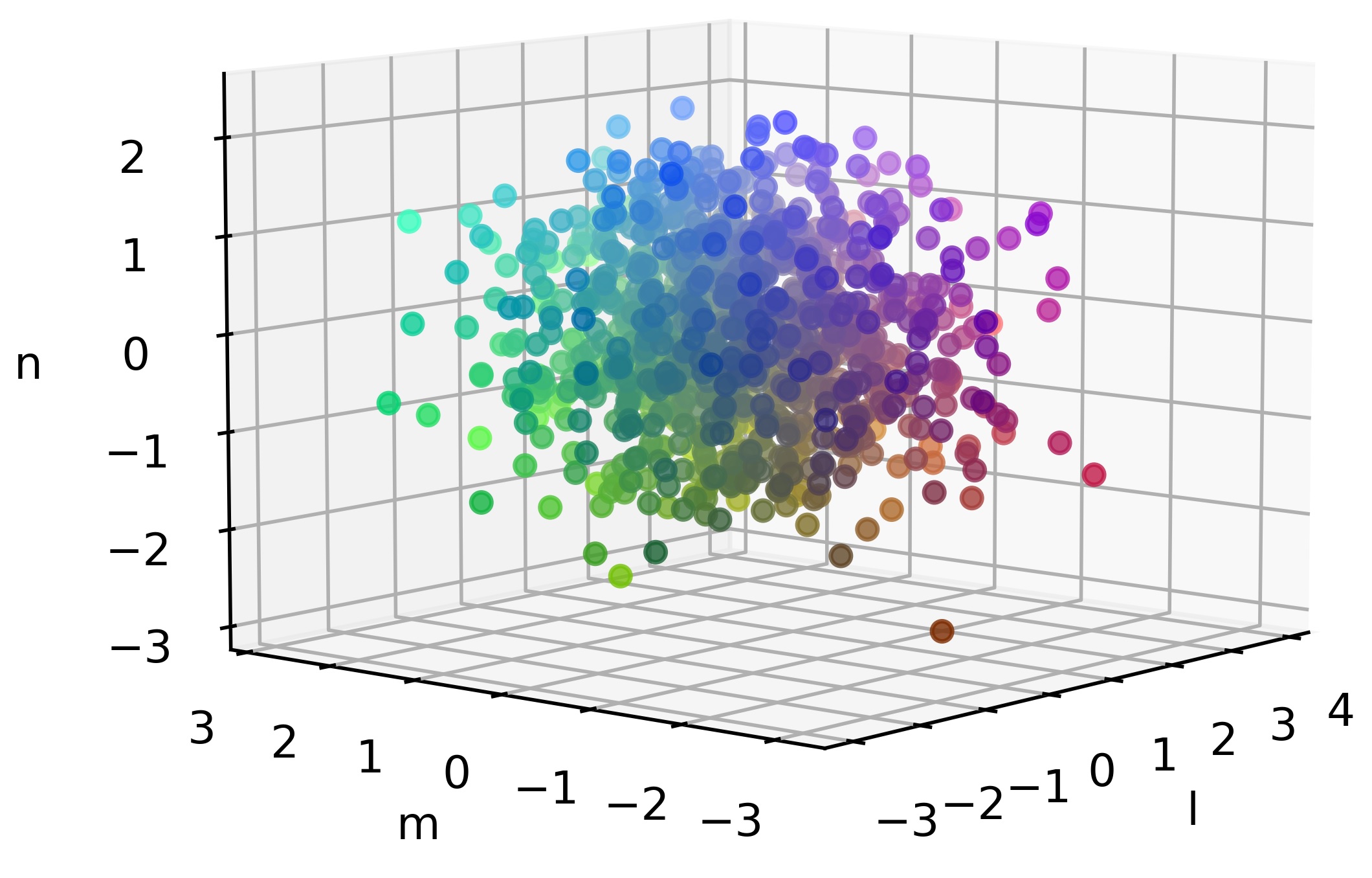}\label{OneEmbeddingAll}}
	\subfigure[Learned action representations with the partition of  500:500]{\includegraphics[width=0.47\columnwidth]{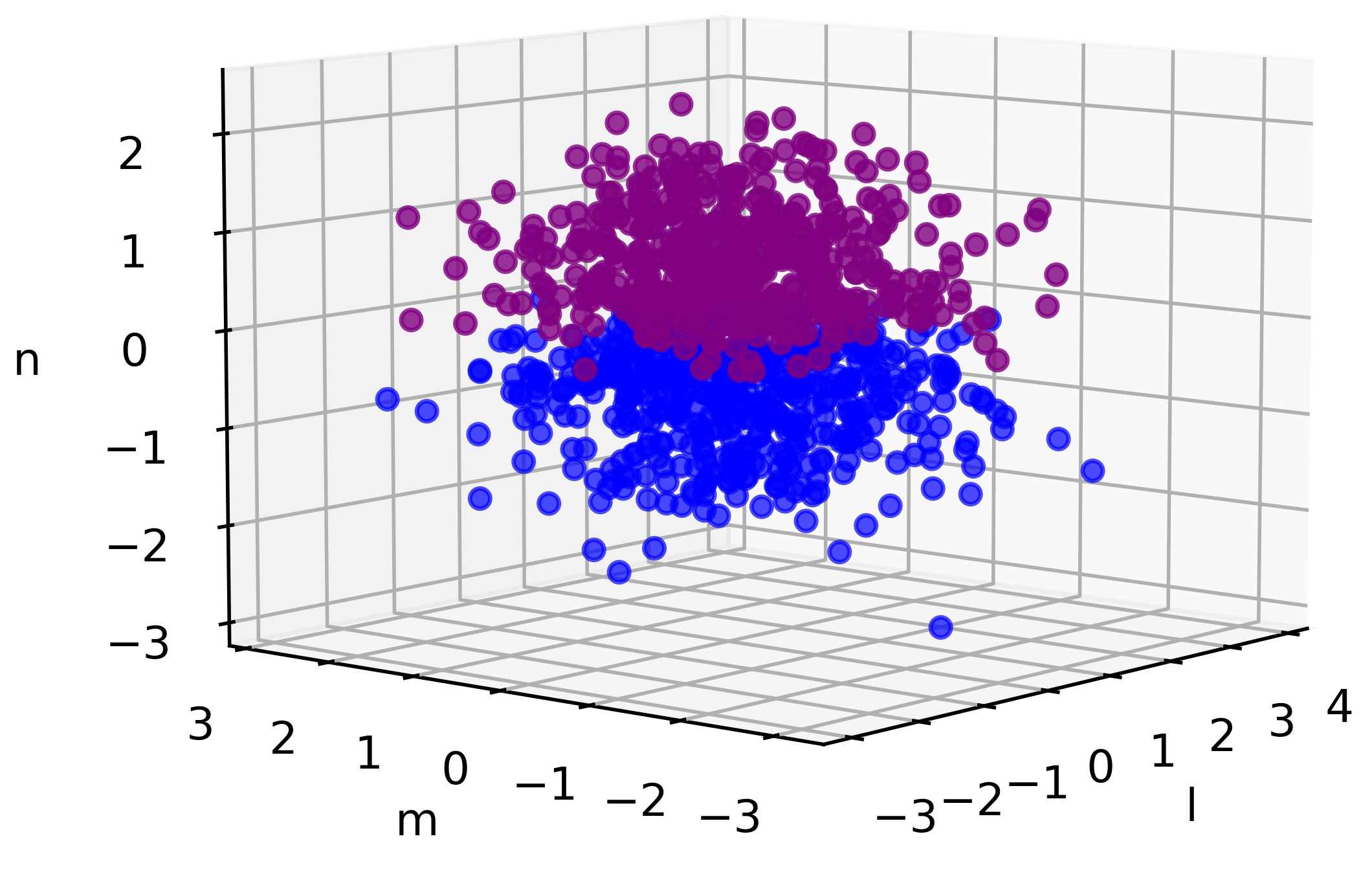}\label{OneEmbedding500}}
	\subfigure[The effect of raw actions on the state transition]{\includegraphics[width=0.47\columnwidth]{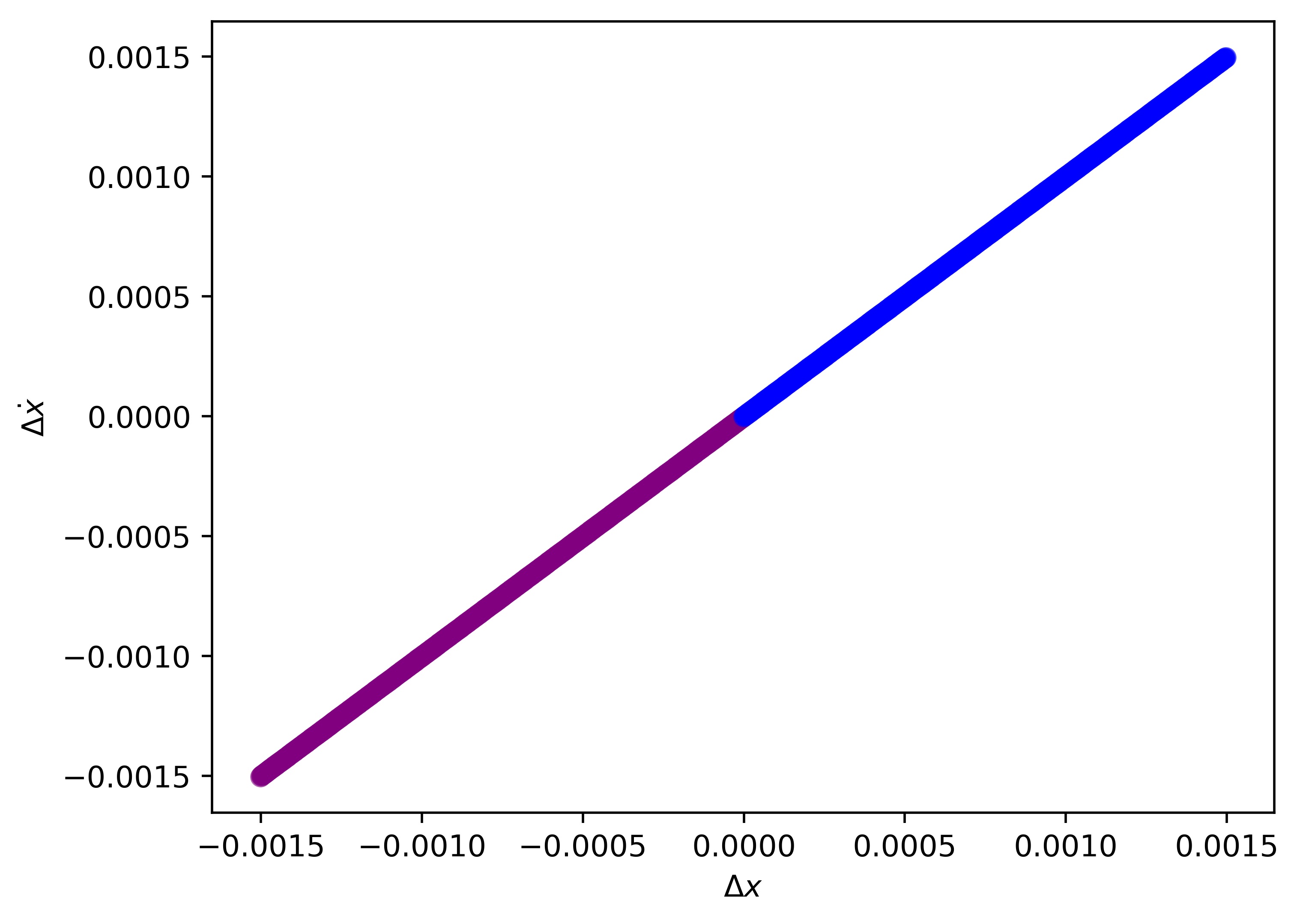}\label{OneRawStateDisplacement}}
	\subfigure[The effect of reconstructed actions on the state transition]{\includegraphics[width=0.47\columnwidth]{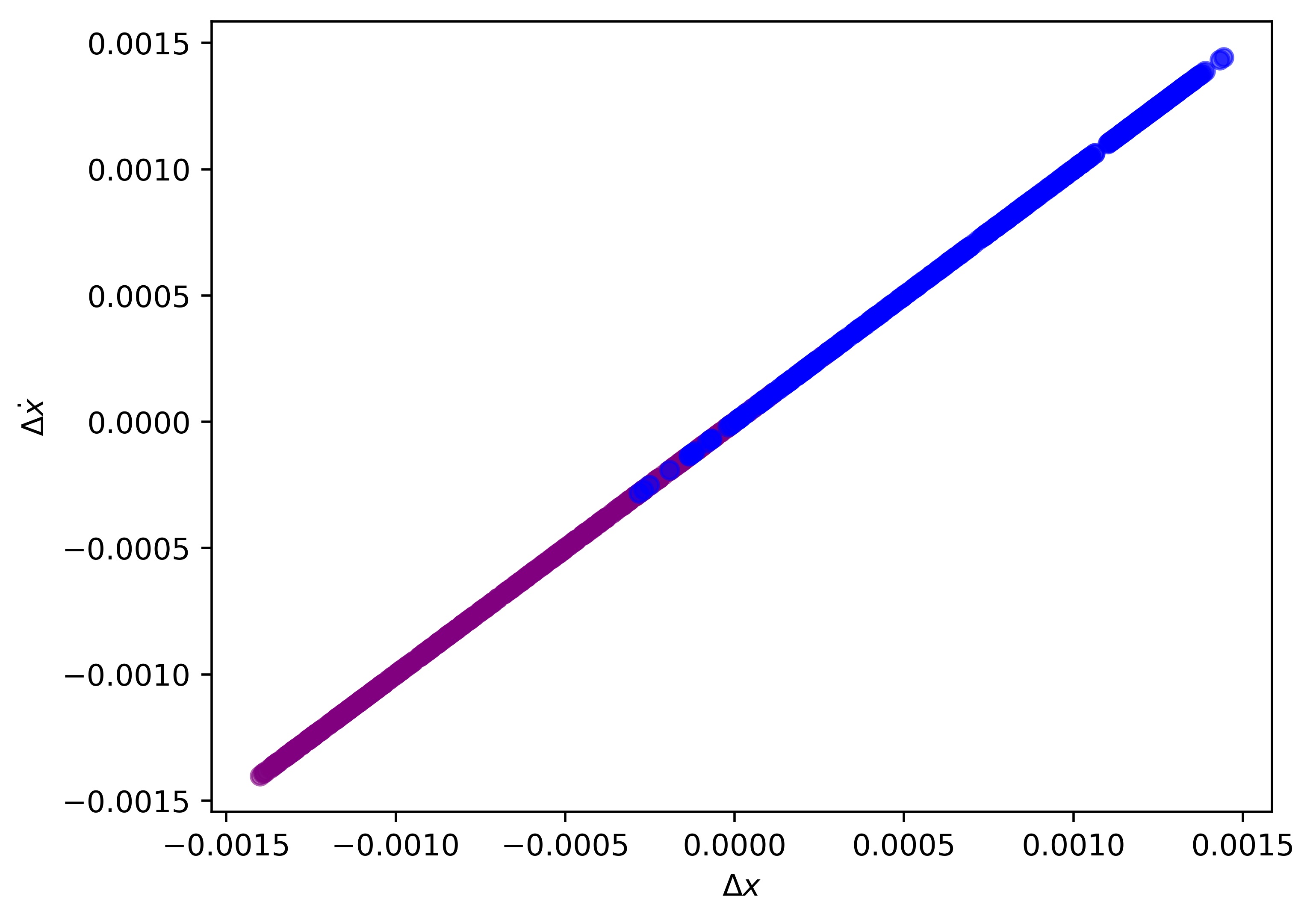}\label{OneReconStateDisplacement}}
	
	\caption{One-step exploration of action representations in latent space. In Fig.8(a) and (b), the $l, m, n$ axes correspond to the three dimensions of the latent space, respectively. In Fig.8(a), 1000 action representations $[e_1, e_2,..., e_{1000}]$ are colored based on the values of each dimension $[l, m, n]$. It can be seen that the color tansitions of the learned action representations are smooth. In Fig.8(b), we take out the action representations $[e_1, e_2,...,e_{500}]$ and mark them as purple, and the action representations $[e_{501}, e_{502},...,e_{1000}]$ are marked as blue. It can be seen that the action representations in latent space are clearly divided into two categories. In Fig.8(c) and (d), the $x, \dot{x}$ correspond to the two dimensions in the state space, i.e., position and velocity, respectively. The comparison of the effects of the raw actions and reconstructed actions on state transitions demonstrates that the reconstructed actions have the similar effect on the state transitions.}
	\label{One-step}
\end{figure}

Then, we investigate the underlying structure and the effect of action representations based on $V_a$ in the scenario of one-step decision making. Here, we set the initial state of the car as $s=( -0.5233, 0)$. Compared with the multi-step task above, the difference lies in the setting of the initial position $x$ of the car. Setting the car at the bottom of the hill is also more challenging. At this given initial position, the car will move to left when the left force is applied; the car will move to right when the right force is applied. Of course, in policy learning, the initial position $x$ of the car is randomly selected from $[-0.6, -0.4]$.  In this experiment, we sample 1000 actions uniformly from the action space, $[a_1, a_2,…, a_{1000}]$, where $\left\{\begin{array}{l}a_{i} \in[-1,0), i=1, \ldots, 500 \\ a_{i} \in(0,1], i=501, \ldots, 1000\end{array}\right.$. For each sampled action, we exert it to the car at the bottom of the hill, and thus get the corresponded 1000 state transitions. Then we use the learned $V_a$ to encode and decode the sampled actions to get the action representations and the reconstructed actions. Finally, the reconstructed actions are used to interact with the environment. In \hyperref[OneEmbeddingAll]{Fig.8(a)}, there are 1000 action representations $[e_1, e_2,..., e_{1000}]$ corresponding to the sampled actions $[a_1, a_2,..., a_{1000}]$, which are colored based on the values of each dimension $[l, m, n]$. It can be seen that the learned action representations are smooth. In \hyperref[OneEmbedding500]{Fig.8(b)}, we take out the action representations $[e_1, e_2,...,e_{500}]$ corresponding to these actions $[a_1, a_2,..., a_{500}]$ and mark them as purple, and take out the action representations $[e_{501}, e_{502},...,e_{1000}]$ corresponding to actions $[a_{501}, a_{501},..., a_{1000}]$ and mark them as blue. It can be seen that the action representations in the latent space are clearly divided into two categories, where one category is the leftward action representations and the other is the rightward action representations. Therefore, the learned action represented latent space well preserves the relative information among all the actions.

In \hyperref[OneRawStateDisplacement]{Fig.8(c)} and \hyperref[OneReconStateDisplacement]{Fig.8(d)}, we compare of the effects of raw actions and reconstructed actions on state transitions. It can be seen that the reconstructed actions obtained from the decoder of $V_a$ have roughly the same effect of the original actions on the state transitions. However, the influence of reconstructed actions on the state transitions are slightly different due to the diversity of samples generated by $V_a$, which can also better explore the state space. This shows that introducing the latent space of action representation can not only learn the relative structures between actions, but also further promote exploration over the state space.

\begin{figure}[h]
	\centering
	\subfigure[Raw actions and reconstructed actions]{\includegraphics[width=0.30\columnwidth]{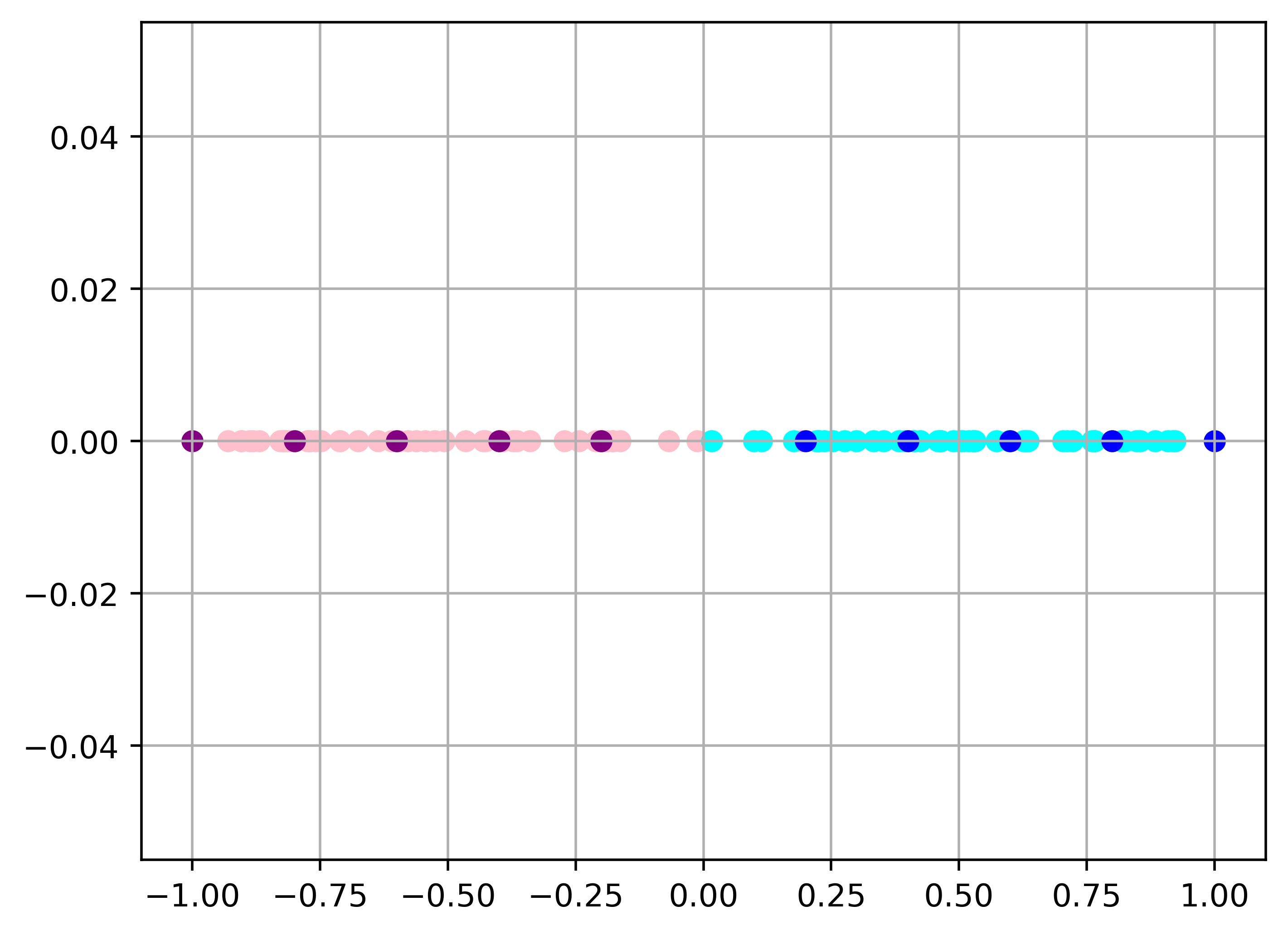}\label{GenExpA}}
	\subfigure[Learned action representations in latent space ]{\includegraphics[width=0.32\columnwidth]{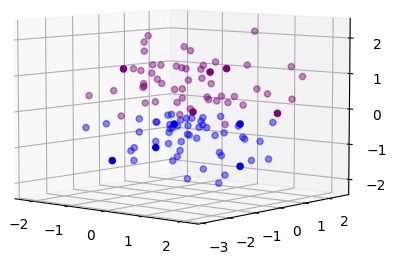}\label{GenExpB}}
	\subfigure[The effect of reconstructed actions on state]{\includegraphics[width=0.32\columnwidth]{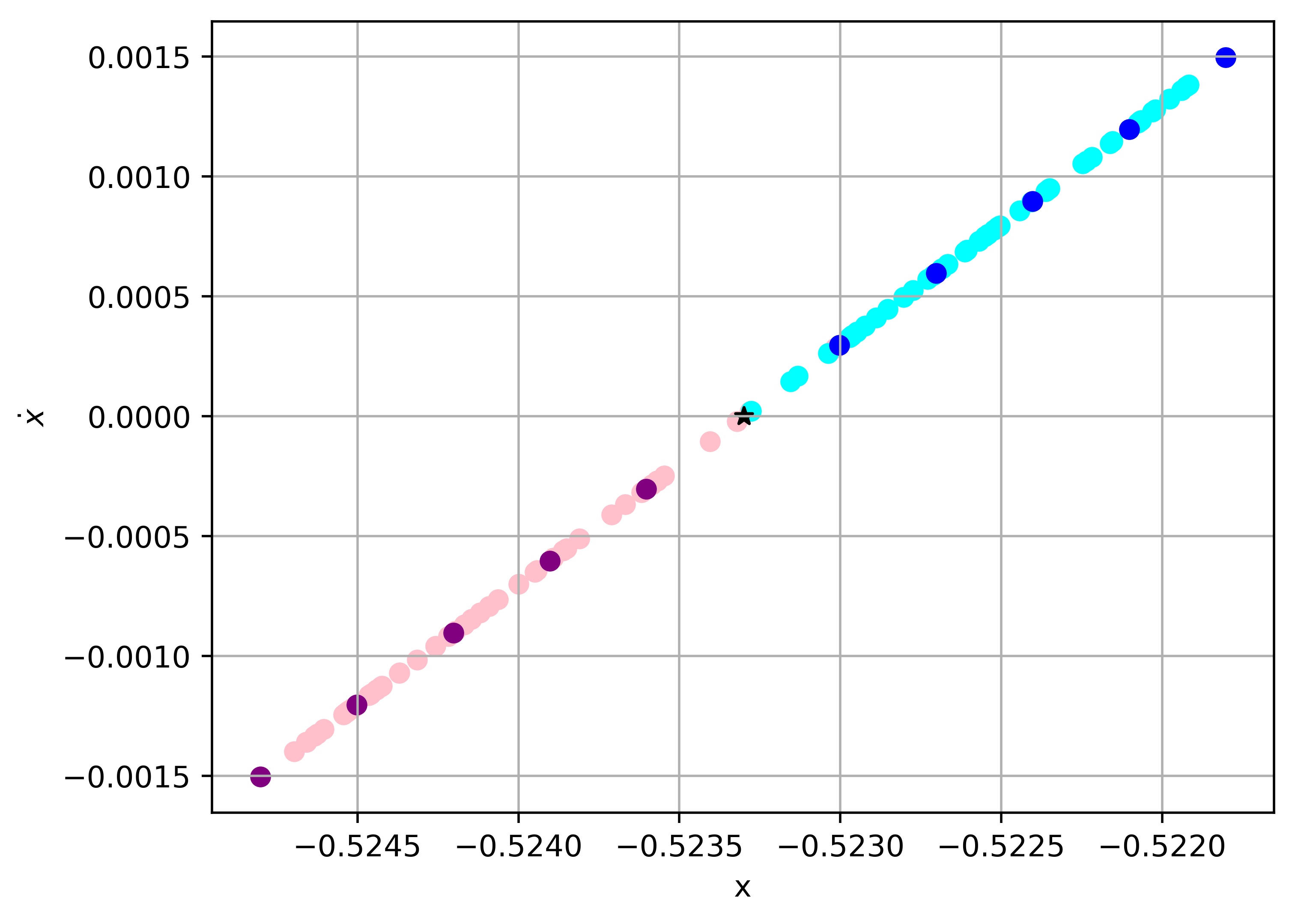}\label{GenExpC}}
	
	\caption{Generalization over actions and exploration over states. In Fig.9(a), the purple points and blue points represent the original actions, and the light purple points and light blue points represent the reconstructed actions obtained by decoding additional sampling of the action representations $e$. Fig.9(b) shows the action representations in latent space. Fig.9(c) shows the effect of interactions with the environment using the reconstructed actions.}
	\label{Generalization}
\end{figure}

\paragraph*{Generalization over Actions and Exploration over States}
The characteristics of sample diversity of VAE has been well demonstrated,
we will investigate whether our proposed action representation model also keep this fine property in DRL tasks. In this experiment, the initial state of the car is set at $s=( -0.5233, 0)$. We sample 5 actions uniformly in intervals $[-1,0)$ and $(0,1]$ in the action space $A$, respectively. Each sampled action is encoded by the learned $V_a$ to obtain its mean $\mu$ and standard deviation $\sigma$ in latent space, and its action representation $e=\mu+\sigma*\varepsilon$ is obtained based on the reparameterization technique with the random variable $\varepsilon \sim N(0,1)$. To investigate the generalization ability, we sample 10 more $\varepsilon$, and generate 10 action representations with mean $\mu$ and standard deviation $\sigma$. With these 10 generated action representations, we get their corresponded reconstructed actions through the decoder of $V_a$. Thus, we generate 10 actions for each raw sampled action in the action space. We plot the raw actions and the reconstructed actions in \hyperref[GenExpA]{Fig.9(a)}, where the dark purple points represent the original actions in  in $[-1,0)$, the dark blue points represent the original actions in $(0,1]$, and the light purple points and light blue points represent the reconstructed actions obtained by decoding the additional sampling of the action representations.
In \hyperref[GenExpB]{Fig.9(b)}, we plot the action representations in latent space, which are divided into two categories obviously, the purple ones correspond to forces to the left direction and the blue points correspond to forces to the right direction. \hyperref[GenExpC]{Fig.9(c)} shows the effect of interactions with environment using the reconstructed actions, which demonstrates that the reconstructed actions enable the agent to explore more unknown states.

Through results shown in \hyperref[Generalization]{Fig.9}, we can see that the latent space of action representations learned by VAE can capture the overall structure of actions in the latent space while also generalizing the action representations in the latent space to other un-executed actions with similar representations, where similar representations refer to similar values of action representations.
Therefore, the proposed VAE-based action representation model significantly improves the generalization performance of action selection, and also encourages explorations over the action and state space.

\paragraph*{Performance Evaluation}

The proposed latent space based policy model is expected to improve the learning efficiency and generalization ability. In this part, we evaluate the performance of the learned policy by our proposed method. To implement the proposed algorithm of policy learning in latent spaces, we
employ the state-of-the-art policy-based algorithm, PPO, as the policy update method. However, we believe that the reported results can be further improved by using our proposed model implemented in other policy learning methods, which is straightforward extention of this work and will leave for our future work.

For the implementation of PPO, we use a neural network framework with shared parameters for the policy and value functions, consisting of a fully connected neural network with two hidden layers, where the number of neurons in the hidden layers is 128 and 64, respectively. The action representation is then mapped to the actual action through the decoder of $V_a$.

To investigate the performance improvement, we compare the following two methods:
\begin{itemize}
	\item \textbf{PPO}: A state-of-the-art policy search method in the field of model-free DRL \citep{PPO2017}.
	\item \textbf{PL-LS with PPO}: Implementation our proposed framework of policy learning in latent spaces with PPO, where the state representation is learned by $V_s$, the action representation is learned by $V_a$.
\end{itemize}

For fair comparison, both compared methods share the same hyperparameters, except for different learning rates. The learning rate of the original PPO method is set to $3e-4$. Since the modules of state representation and action representation in latent spaces are included in the framework of PL-LS with PPO, the overall structure of the compared methods are different. Thus, it is not necessary for the learning rates to be exactly the same. The detailed parameter settings are concluded in \hyperref[table:MountainCarParamter]{Table 1}.

\begin{table}[h]
	\centering
	\caption{Hyperparameters of PL-LS with PPO in MountainCar task.}
	\label{table:MountainCarParamter}
	\small
	\begin{tabular}{|c|c|}
		\hline
		Hyperparameter             & Value \\ \hline
		Horizon                    & 512   \\ \hline
		Learning rate (Adam)       & 4e-4  \\ \hline
		Num. epochs                & 10    \\ \hline
		Minibatch size             & 128   \\ \hline
		Num. parallel environments & 32    \\ \hline
		Discount ($\gamma$)          & 0.99  \\ \hline
		GAE parameter ($\lambda$)    & 0.95  \\ \hline
		Clipping parameter $\epsilon$ & 0.2   \\ \hline
		VF coeff. $c_{1}$            & 0.5   \\ \hline
		Entropy coeff. $c_{2}$       & 0.01  \\ \hline
	\end{tabular}
\end{table}

\begin{figure}[H]
	\centering
	\includegraphics[scale=.64]{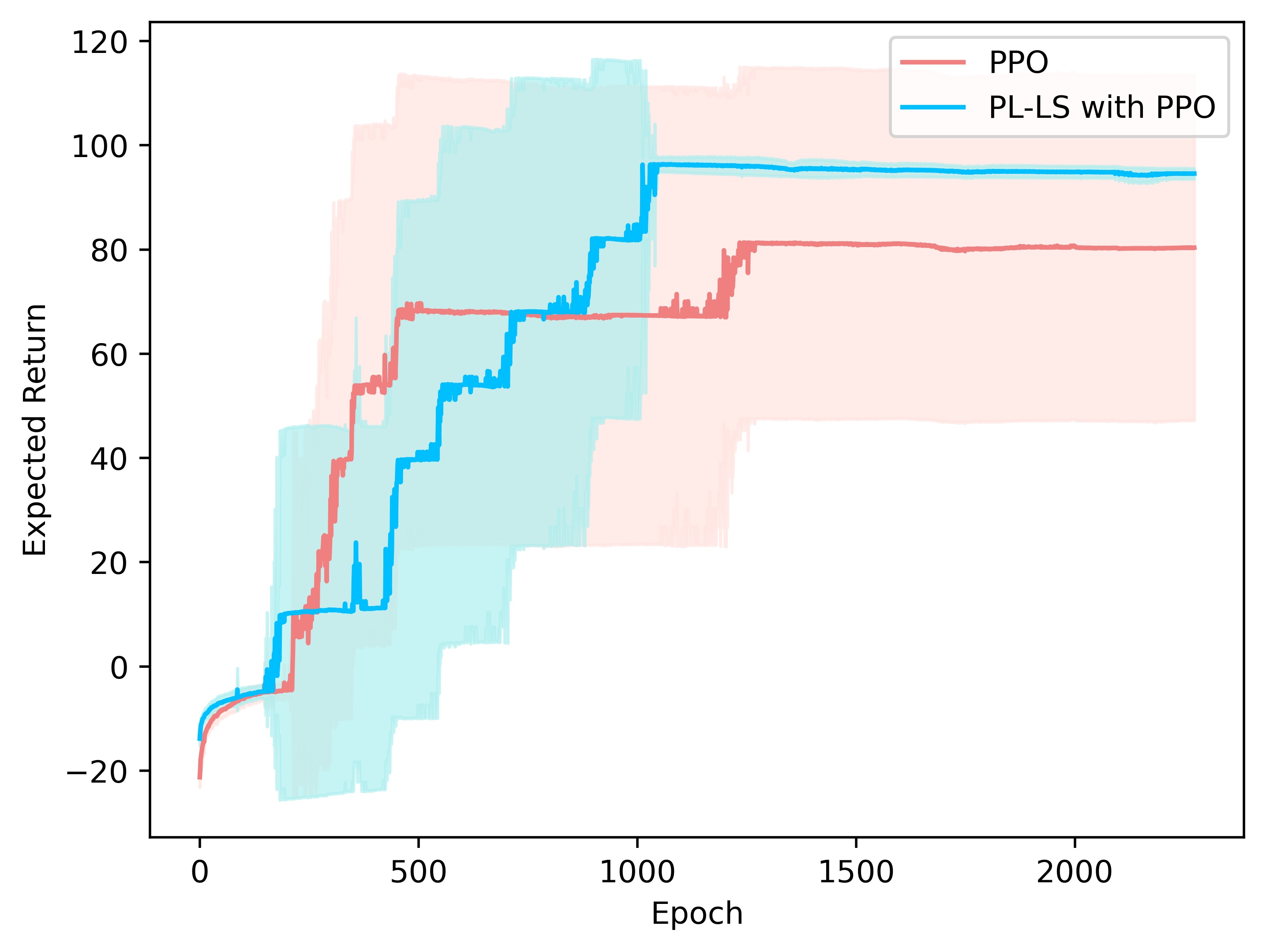}
	\caption{Average expected return over 7 trials for MountainCar. The horizontal coordinate indicates the number of epochs, and the vertical coordinate indicates the obtained expected return. The shaded areas are the standard deviations.} \label{MountainCarReturn}
\end{figure}

We investigate the average expected return over 7 trials for the MountainCar task, and each trial is with a different random seed. In each trial, the expected return is calculated over 10 completely new testing episodes with the same random seed. The experimental results are plotted in \hyperref[MountainCarReturn]{Fig.10}, which show that our proposed method PL-LS with PPO obtains performs better than the original PPO method, especially when the Epoch reaches 750. The method of PL-LS with PPO converges faster, and with a small standard deviation after $Epoch=1000$, while the original PPO method starts to converge after $Epoch=1200$, but it still with a large standard deviation. This shows that the proposed PL-LS with PPO learns the policy more stable and is with better robustness; While the original PPO method is slightly inferior. Note that the large standard deviation in the initial stage of learning (e.g., $Epoch < 1000$) is due to its sparse reward in this task, because the reward is always negative at the beginning and the car is rarely guided by a positive reward before the car reaches the hilltop goal.

\begin{figure}[H]
	\centering
	\includegraphics[scale=.41]{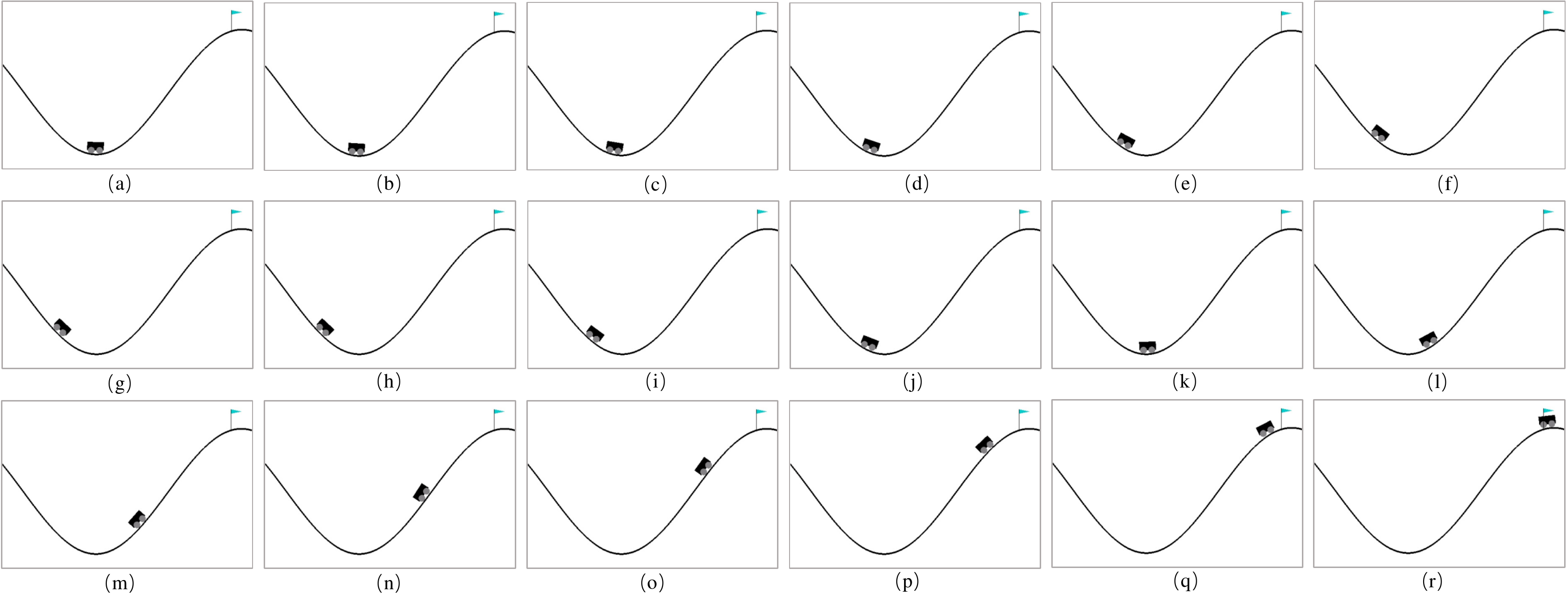}
	\caption{The demonstration of learned policy by PL-LS, where the sampling interval is 4 time steps.} \label{MountainCarRun}
\end{figure}

Finally, we illustrate an example of the policy learned through PL-LS with PPO. The trajectory is drawn in \hyperref[MountainCarRun]{Fig.11}, where the sampling interval is 4 time steps. In this illustration, the initial state of the car is set to $s=(-0.5309, 0.)$, as shown in \hyperref[MountainCarRun]{Fig.11(a)}, and the goal is to move the car to the location of the flag, as shown in \hyperref[MountainCarRun]{Fig.11(r)}. \hyperref[MountainCarRun]{Fig.11(a)}-\hyperref[MountainCarRun]{Fig.11(g)} shows that the car is moving to the left to gain power. \hyperref[MountainCarRun]{Fig.11(h)}-\hyperref[MountainCarRun]{Fig.11(r)} shows that the car is moving to the right, climbing up the hill to reach the target position. With the learned policy, the car takes 68 time steps to reach the target location from the initial position. It can be seen that the proposed method can accomplish the task of MountainCar very well.

\subsection{CarRacing}

In this section, we evaluate the performance of our proposed model on a more challenging task, CarRacing. In this task, the car drives on a randomly generated track in each trial, and its goal is to visit as many tiles as possible within the least amount of time so as to obtain higher rewards. The state is composed of classic RGB images with 96*96 pixels, which is viewed from a top-down angle. Each pixel is stored with three floating point values between 0 and 1. The action space is three-dimensional and continuous, corresponding to the steering angle $o \in [-1,1]$, the acceleration $p \in [0,1]$ and the brake $q \in [0,1]$ respectively, i.e., $a=(o,p,q)$. The reward is equal to -0.1 every frame and +1000/N for every track tile visited, where N is represented by the total number of tiles in track.

In the CarRacing task, we verify the effectiveness and applicability of introducing both state and action representations, and investigate the impact of policy learning in latent spaces.

\paragraph*{The State Representations in Latent Space} In order to obtain the representation of states in latent space, we use the model of VAE for training. We collect data of 57 trajectories with a random policy, which includes 57000 state data. We use 54,000 data as the training set and 3,000 data as the test set.

\begin{figure}[h]
	\centering
	\includegraphics[scale=.69]{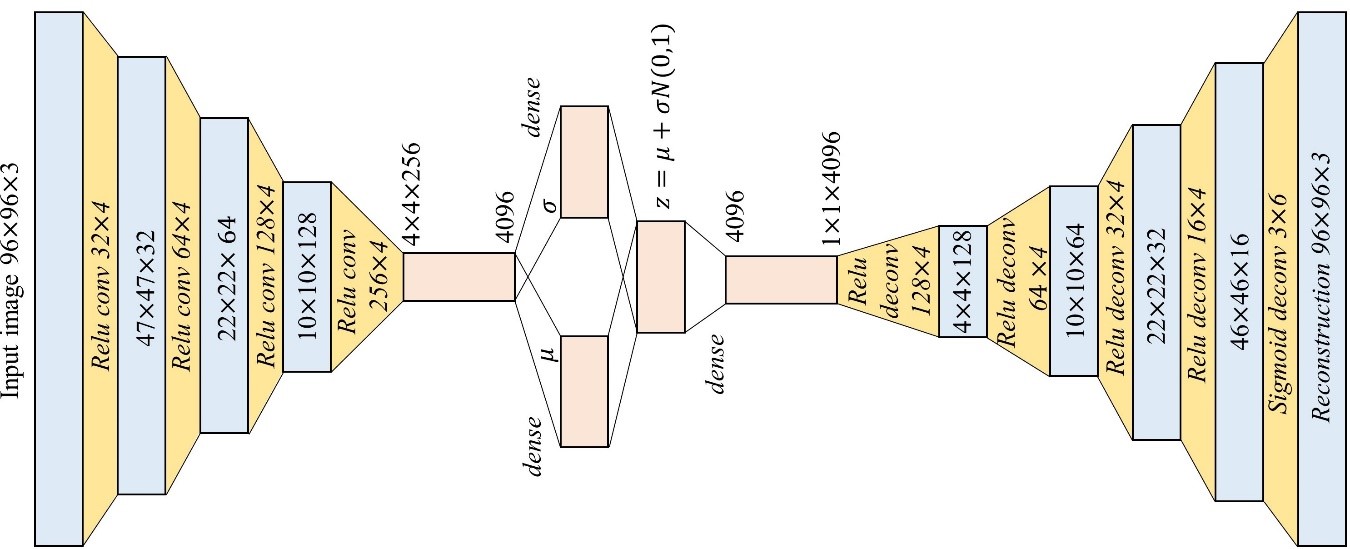}
	\caption{Structure of $V_s$ used to train the state representations in latent space. The encoder contains four convolutional layers and the decoder contains four deconvolutional layers. The activation functions are marked in Italics.} \label{StateVae}
\end{figure}

\begin{figure}[h]
	\centering
	\includegraphics[scale=.60]{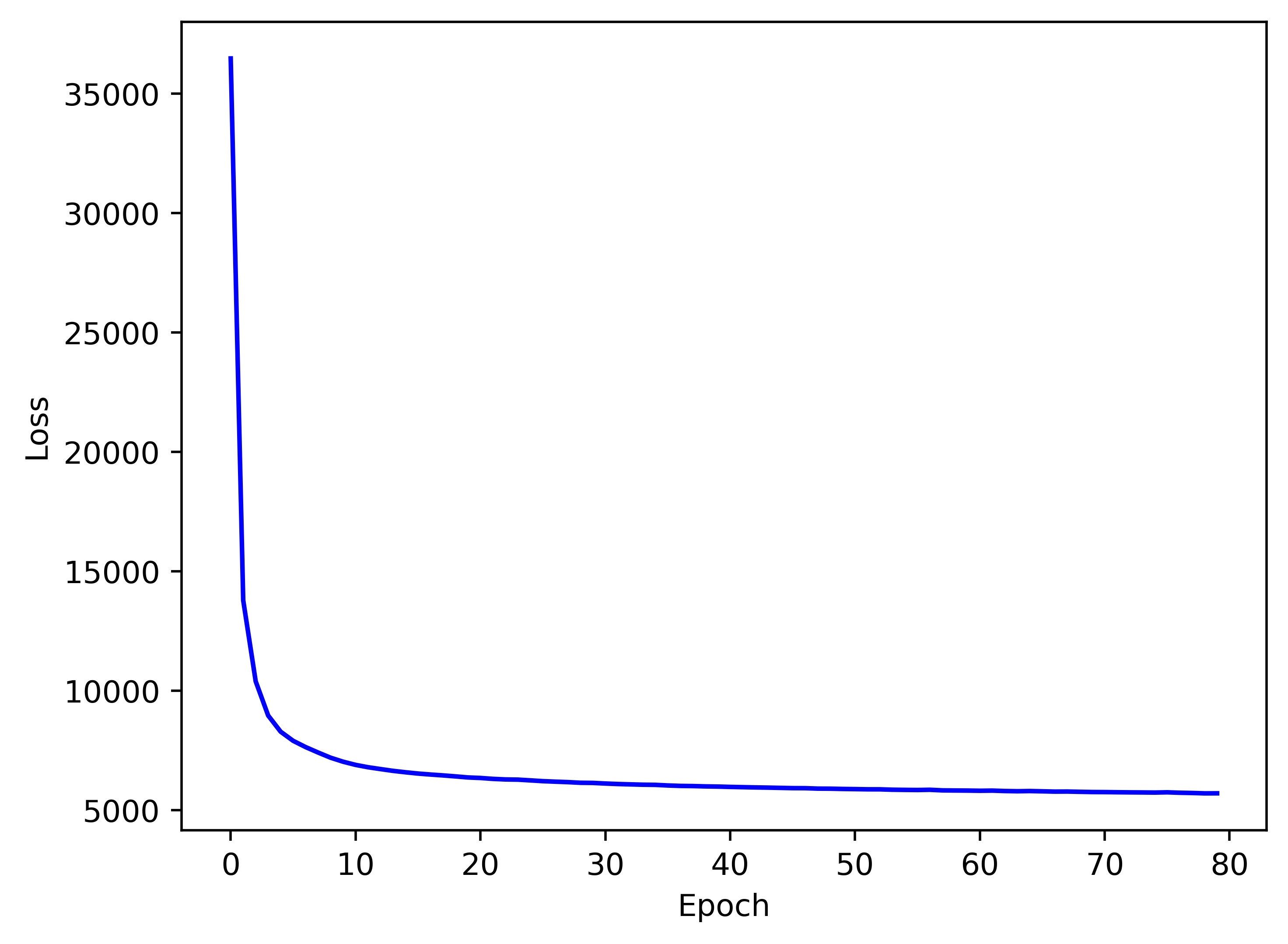}
	\caption{Training process of $V_s$. The horizontal coordinate indicates the number of epochs during training, and the vertical coordinate indicates the loss during training.} \label{trainStatelossCarRacing}
\end{figure}

The structure of the state representation model VAE($s_t$), abbreviated as $V_s$, is shown in \hyperref[StateVae]{Fig.12}. The input of $V_s$ is a 96*96*3 tensor, and is passed through 4 convolutional layers and encoded into 32-dimensional latent mean vector $\mu$ and deviation vector $\sigma$, each of size is $N_z$=32. The dimensionality of the latent vector, $N_z=32$, is choosen with reference to the one used in World Model \cite{worldmodels}. The latent vector $z$ is sampled from the Gaussian distribution $N(\mu, \sigma I)$. The latent vector $z$ is then passed through the deconvolution layer to decode and reconstruct the image. Each convolution and deconvolution layer has a stride of 2. These layers are colored by yellow in the diagram, which contain the activation function and the size of the filter. All convolution and deconvolution layers use the Relu activation function, except for the output layer, as we need the output to be between 0 and 1. We use mini-batch stochastic gradient descent with Adam optimizer to train $V_s$. The learning rate is set as $0.0001$. The training process of $V_s$ is shown in \hyperref[trainStatelossCarRacing]{Fig.13}.


\begin{figure}[H]
	\centering
	\subfigure[Original observed state (Car is on the driveway)]{\includegraphics[width=0.24\columnwidth]{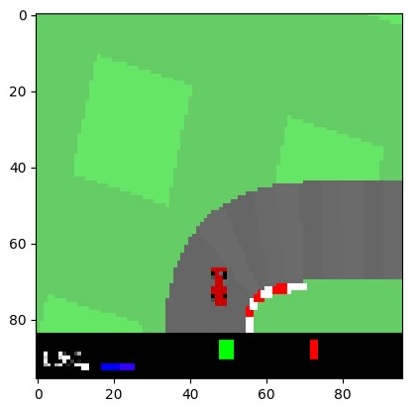}\label{RawCarRacing1}}
	\subfigure[Reconstructed state (Car is on the driveway)]{\includegraphics[width=0.24\columnwidth]{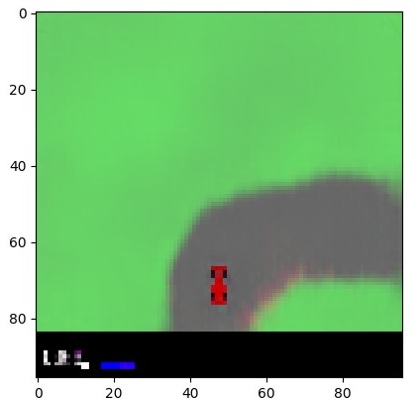}\label{ReconCarRacing1}}
	\subfigure[Original observed state (Car is on the grass)]{\includegraphics[width=0.24\columnwidth]{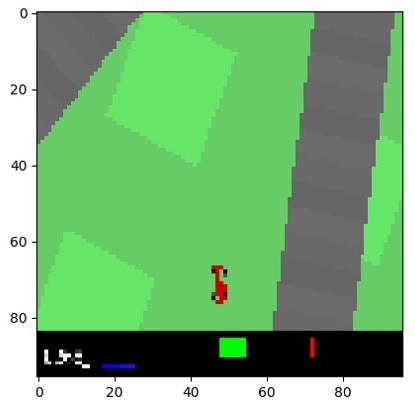}\label{RawCarRacing2}}
	\subfigure[Reconstructed state (Car is on the grass)]{\includegraphics[width=0.24\columnwidth]{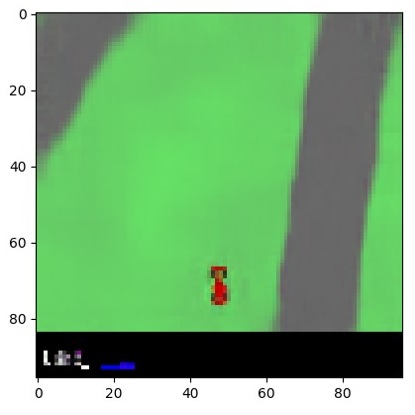}\label{ReconCarRacing2}}
	
	\caption{Examples of reconstructed states, which are obtained by the decoder of the final converged state representation model $V_s$.}
	\label{CarRacing}
\end{figure}

When the $V_s$ model converges, we test its performance with the test set and observe the reconstructed images as shown in \hyperref[CarRacing]{Fig.14}. \hyperref[RawCarRacing1]{Fig.14(a)} and \hyperref[RawCarRacing2]{(c)} show the original images with different car position,  and \hyperref[ReconCarRacing1]{Fig.14(b)} and \hyperref[ReconCarRacing2]{(d)} show the reconstructed images. It can be seen that $V_s$ is able to reconstruct the state images very well, whether the car is on the driveway or the car is on the grass.

\begin{figure}[H]
	\centering
	\includegraphics[scale=.60]{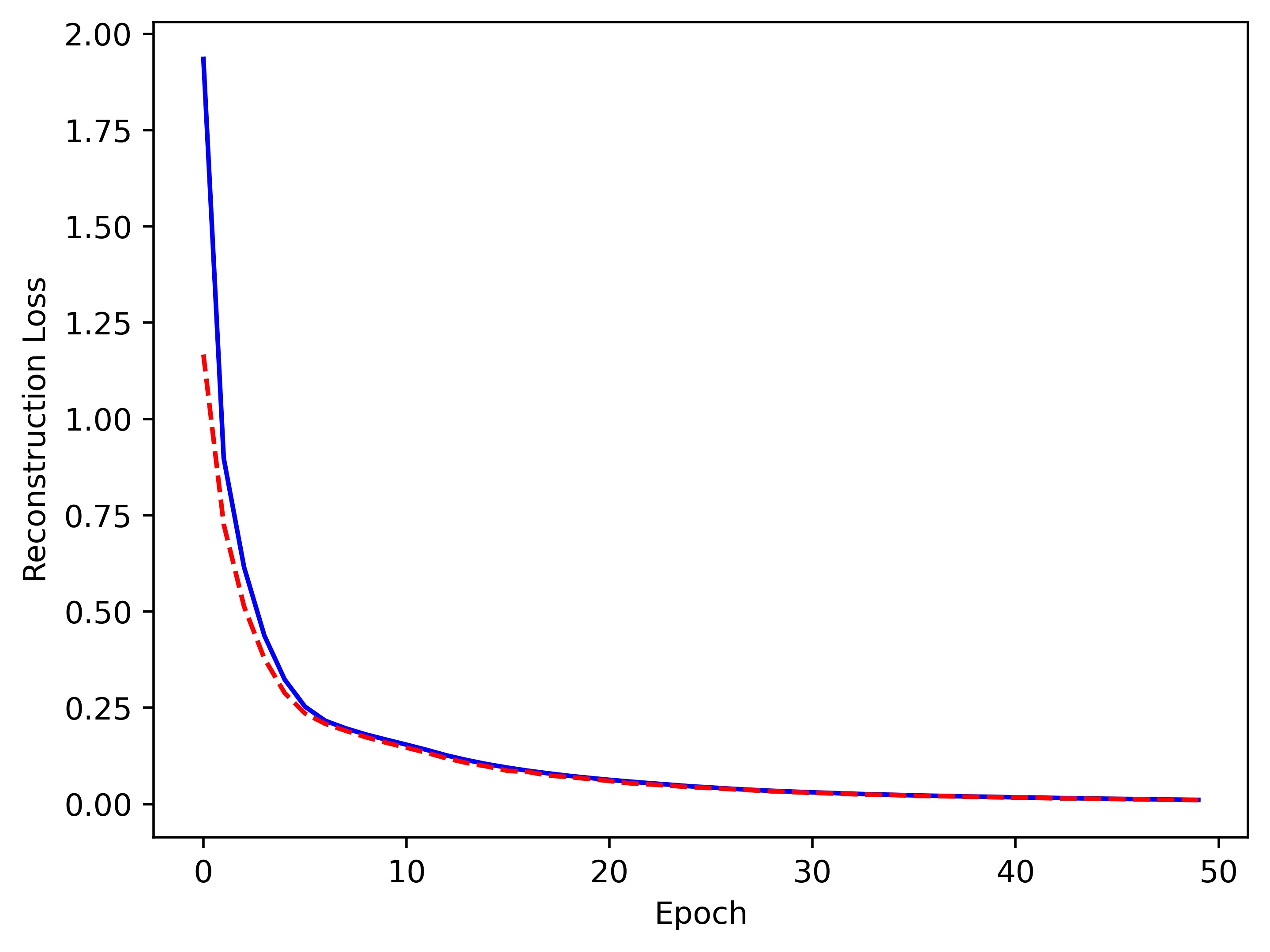}
	\caption{Learning process of $V_a$. The horizontal coordinate indicates the number of epochs during training, and the vertical coordinate indicates the reconstruction loss. The blue line indicates the reconstruction loss during training and the red line indicates the reconstruction loss during validation.} \label{trainActionlossCarRacing}
\end{figure}

\paragraph*{The Action Representations in Latent Space} Similarly, we use VAE($a_t$), abbreviated as $V_a$, to learn the action representations in latent space. The training is performed using the action data in the trajectories collected above. The 3-dimensional action data is taken as the input of $V_a$ and encoded into latent mean vector $\mu'$ and latent deviation vector $\sigma'$ by a layer of full connections with 10 units, and then we sample the latent vector $e$ from the Gaussian distribution $N(\mu', \sigma' I)$. Based on our preliminary experiments, the dimensionality of the action representation $e$ is set as $N_e$=32, which performs the best.
The action representation $e$ is passed through a layer of full connections with 10 units and finally reach the output layer to get the reconstructed action.
We use the same traning strategy as $V_s$ to train $V_a$, but with a learning rate of $0.0001$. The learning process of $V_a$ is shown in \hyperref[trainActionlossCarRacing]{Fig.15}.

\begin{figure}[h]
	\centering
	\includegraphics[scale=.73]{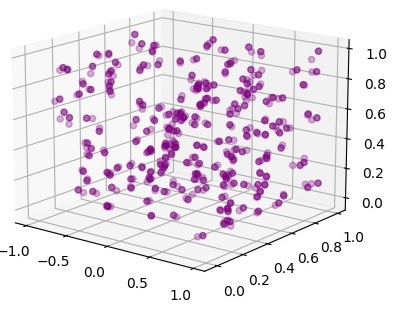}
	\caption{Original actions and corresponding reconstructed actions by $V_a$. The original actions are colored in dark purple and the reconstructed actions are colored in light purple.} \label{ActionVAE}
\end{figure}

We evaluate the performance of the reconstructed action by MSE, which is $0.00092 \pm (7.39\times10^{-6})$. We also draw the first 200 of the test data and the corresponding reconstructed actions, as shown in \hyperref[ActionVAE]{Fig.16}. The original actions are colored in dark purple and the reconstructed actions are colored in light purple. The results show that the action representations in latent space can generalize the action to more unseen actions with similar representations, and thus the action and state spaces are better explored.


\paragraph*{Policy Learning} To learn the policy of this CarRacing task, we also implement our proposed latent spaces based policy learning method with PPO algorithm. In this part, we compare the following two methods:
\begin{itemize}
	\item \textbf{PPO}: The plain PPO method \citep{PPO2017}.
	\item \textbf{PL-LS with PPO}: 	Implementation our proposed framework of policy learning in latent spaces with PPO, where the state representation is learned by $V_s$, the action representation is learned by $V_a$.
\end{itemize}

\begin{figure}[h]
	\centering
	\subfigure[The overall structure of PL-LS with PPO ]{\includegraphics[width=0.83\columnwidth]{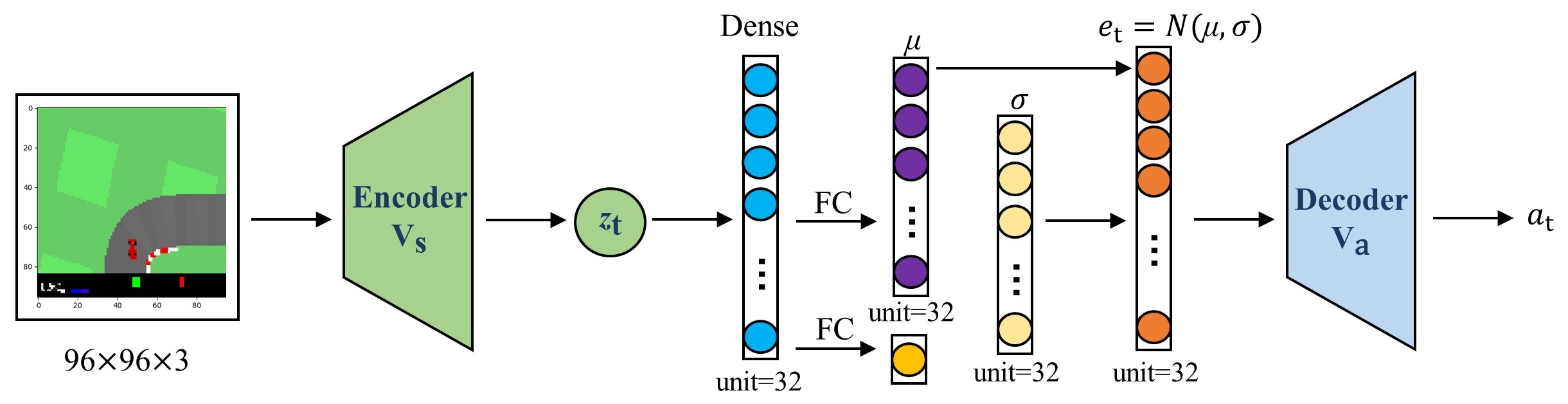}\label{VPPO}} \\
	\subfigure[The structure of policy model in plain PPO]{\includegraphics[width=0.83\columnwidth]{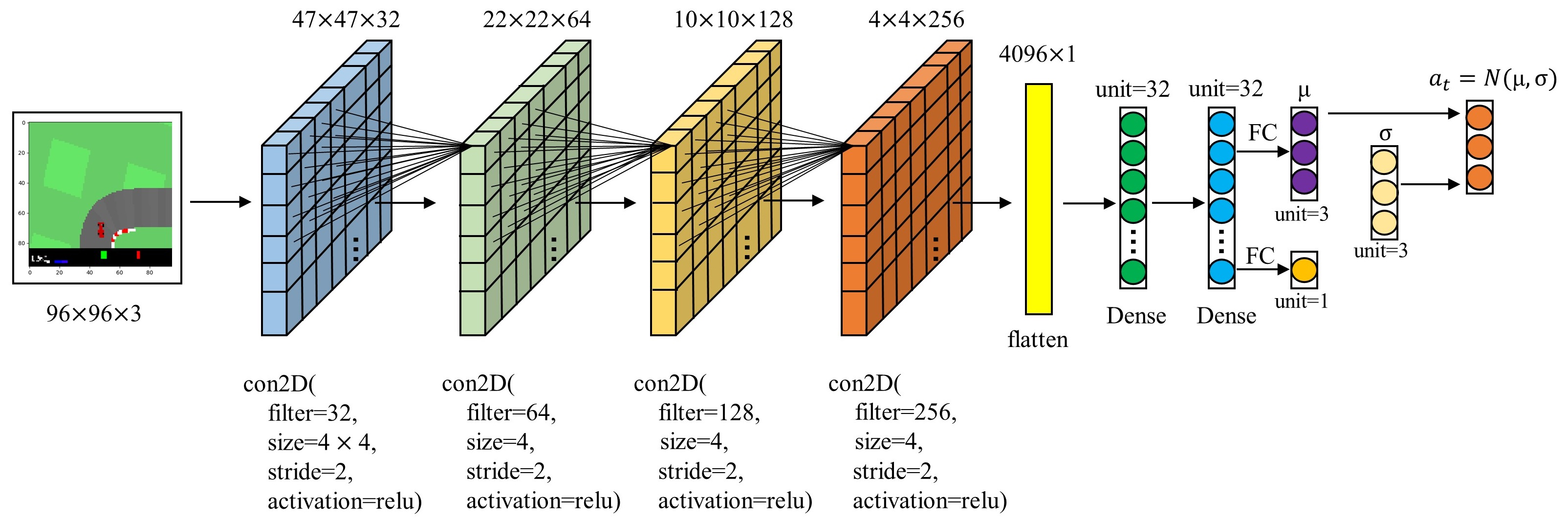}\label{PPO_1}}
	\caption{The comparison of network structure between the two methods. In the proposed PL-LS with PPO method, only the parameters of the policy model are constantly being trained and updated in the online manner. In plain PPO method, all parameters of the large-scale network are constantly being trained and updated.}
	\label{ComparisonNetwork}
\end{figure}

The network structures of the two methods are shown in \hyperref[ComparisonNetwork]{Fig.17}.
In the implementation of our proposed PL-LS, we input the observed image into the encoder of $V_s$ to obtain the state representation in latent space. The policy model is implemented by a neural network with shared parameters for the policy and value functions, which is composed of a fully connected neural network, including one hidden layer with 32 units. The state representation is inputed to the policy model, and the action representation is obtained. The action representation in latent space is then decoded to the actual action by the decoder of $V_a$, as shown in \hyperref[VPPO]{Fig.17(a)}.
The network structure of plain PPO contains four convolutional layers and two fully connected layers, and all the network parameters need to be updated in each iteration, as shown in \hyperref[PPO_1]{Fig.17(b)}. To be fair, we try to make the network structure at each component correspondingly to be the same for the compared methods.

Both these two compared methods use the same parameters, but the learning rate of PPO is fixed. The detailed parameter settings are shown in \hyperref[table:CarRacingParamter]{Table 2}.

\begin{table}[H]
	\centering
	\caption{Hyperparameters of PL-LS with PPO used in CarRacing.}
	\label{table:CarRacingParamter}
	\small
	\begin{tabular}{|c|c|}
		\hline
		Hyperparameter             & Value \\ \hline
		Horizon                    & 1000   \\ \hline
		Learning rate (Adam)       & 1e-4  \\ \hline
		Num. epochs                & 10    \\ \hline
		Minibatch size             & 1000   \\ \hline
		Num. parallel environments & 16    \\ \hline
		Discount ($\gamma$)          & 0.99  \\ \hline
		GAE parameter ($\lambda$)    & 0.95  \\ \hline
		Clipping parameter $\epsilon$ & 0.2   \\ \hline
		VF coeff. $c_{1}$            & 0.5   \\ \hline
		Entropy coeff. $c_{2}$       & 0.01  \\ \hline
	\end{tabular}
\end{table}

\begin{figure}[H]
	\centering
	\includegraphics[scale=.6]{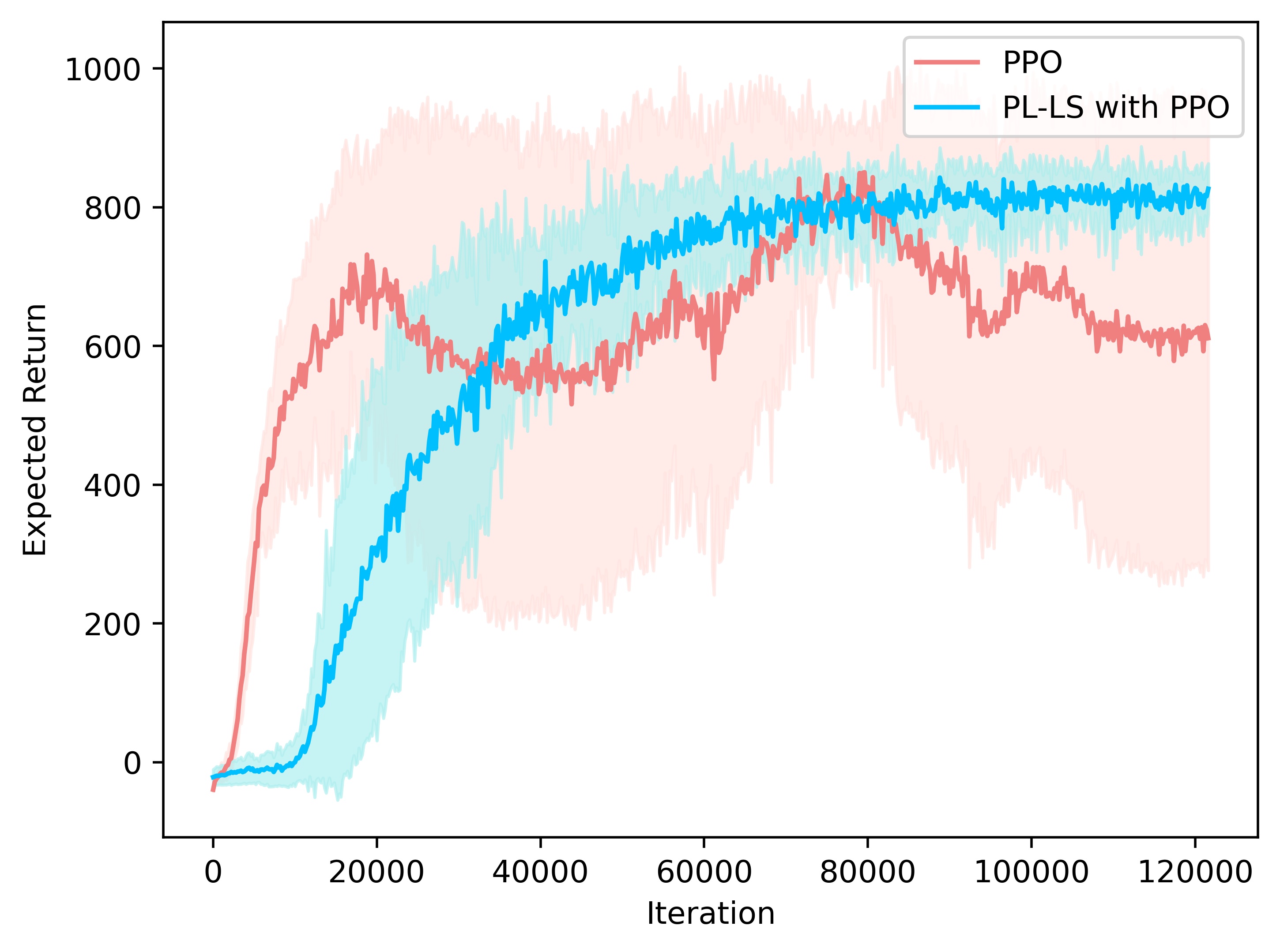}
	\caption{Average expected return over 10 trials for CarRacing. The horizontal coordinate indicates the number of model iterations and the vertical coordinate indicates the expected return. The shaded areas are the standard deviations.} \label{CarRacingReturn}
\end{figure}

We investigate the average return for CarRacing over 10 trials. In each trial, the expected return is calculated over 10 test episodic samples (which are not used for policy learning). The experimental results are plotted in \hyperref[CarRacingReturn]{Fig.18}, which show that PL-LS with PPO outperforms the plain PPO method. The expected reward of the proposed PL-LS with PPO method is increasing stably and eventually converged. The return of the plain PPO method, on the other hand, is more curvilinear and the variance seems to be severer.

On the other hand, we investigate the learning efficiency by summarizing the total number of parameters, the number of trainable parameters involved in the models during training process, and the frequency of samples collected by the agent from the 16 parallel environments before the policies get converged as shown in \hyperref[table:Params]{Table 3}, where the policy is converged at 60,000 iterations for the method of PL-LS with PPO, while the plain PPO method still does not show significant convergence at 120,000 iterations according to the results in \hyperref[CarRacingReturn]{Fig.18}.
It can be seen that our proposed latent spaces based policy model is more compact, the network scale is more lightweight, and required sample size for policy learning is less. Therefore, the learning efficiency of the proposed method is promising.

\begin{table}[H]
	\centering
	\caption{Comparisons in terms of learning efficiency}
	\label{table:Params}
	\small
	\resizebox{\textwidth}{!}{
	\begin{tabular}{|c|c|c|c|}
		\hline
		Model          & Total\_Parameters & Trainable\_Parameters & Sampled\_Frequency \\ \hline
		PL-LS with PPO & 8710              & 4354                  & 3750               \\ \hline
		PPO            & 3289768           & 1644878               & 7500               \\ \hline
	\end{tabular}
    }
\end{table}

\begin{figure}[H]
	\centering
	\includegraphics[scale=.41]{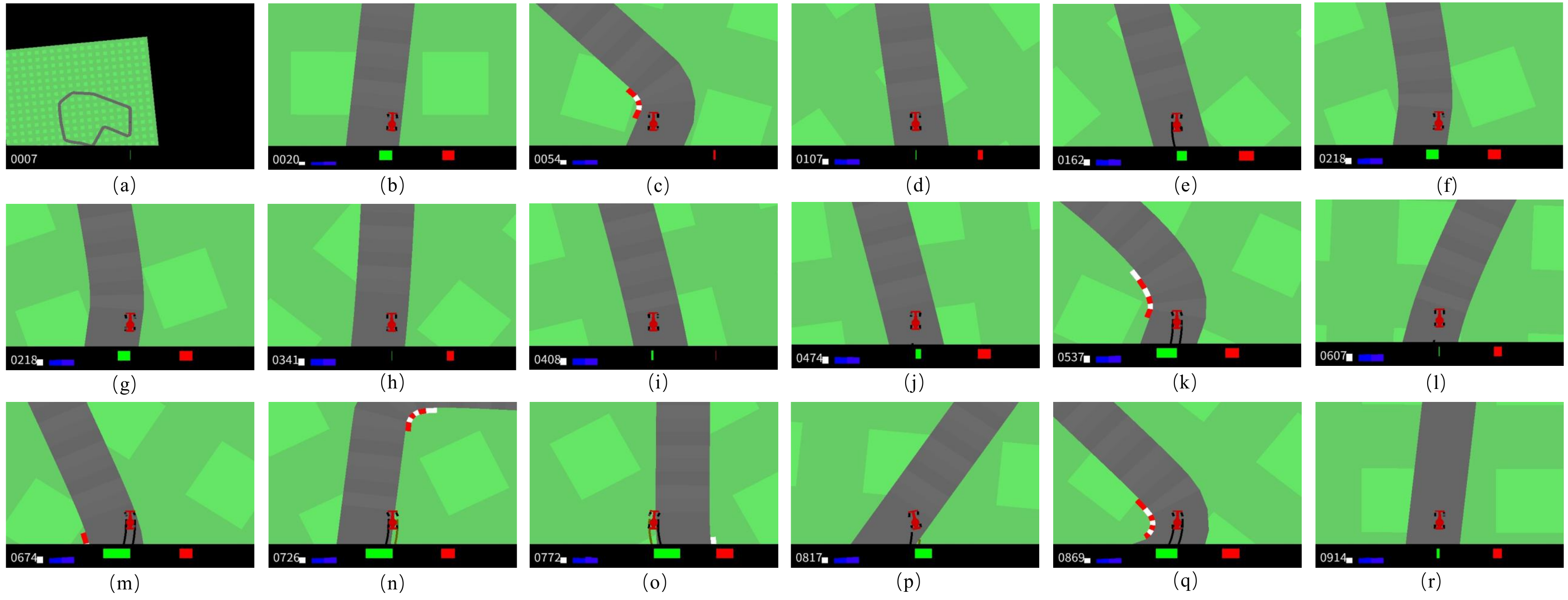}
	\caption{The demonstration of learned policy by PL-LS. The sampling interval is 48 frames.} \label{CarRacing_18}
\end{figure}

At last, we illustrate how the learned policy by the proposed method PL-LS with PPO performs. The trajectory is drawn in \hyperref[CarRacing_18]{Fig.19}, where the sampling interval is 48 frames. In this illustration, we can see that the car can drive smoothly on the track, and eventually complete the entire track quickly. With the policy learned by PL-LS with PPO, the car can return to the track even if the car drives to the grass. However, when the car drives to the grass, it always spins on the grass and is difficult to go back to the track with the policy learned by the plain PPO method. It can be seen that the PL-LS method can efficiently complete the task of CarRacing.

\begin{figure}[h]
	\centering
	\includegraphics[scale=.67]{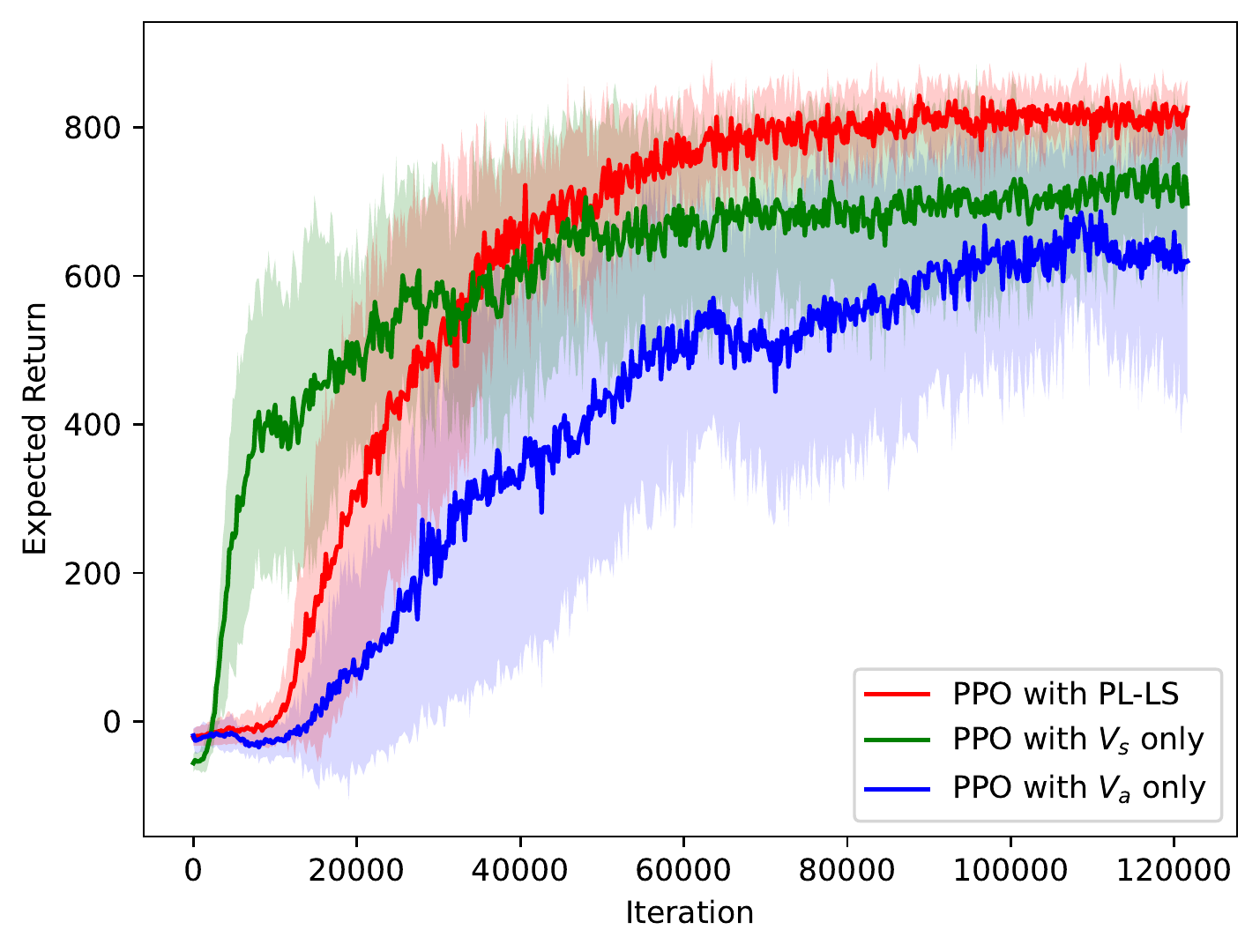}
	\caption{Ablation study in terms of policy performance over 10 trials for CarRacing. The horizontal coordinate indicates the number of learning iterations and the vertical coordinate indicates the expected return. The shaded areas are the standard deviations.} \label{Ablation}
\end{figure}

\paragraph*{Ablation Study} Finally, we conduct ablation experiment to understand the impact of each component in the PL-LS framework. We first analyze three components in terms of the model framework, including $V_s$, which learns the state representations in latent space; $V_a$, which learns the action representations in latent space; The policy learning method. In \hyperref[Ablation]{Fig.20}, we show the performance of the ablation study, where the green curve indicates a policy learning method with state representation $V_s$ only (i.e., PPO with $V_s$ only), the blue curve indicates a policy learning method with action representation $V_a$ only (i.e., PPO with $V_a$ only), and the red curve indicates a policy learning method with both state representation $V_s$ and action representation $V_a$ (i.e., PPO with PL-LS). We can see that the performance of the policy works best when both $V_s$ and $V_a$ are included, indicating that learning the state and action representations in latent space can further improve the learning performance of the policy.

\subsection{Cheetah}
In this section, we implement the proposed framework PL-LS with the algorithm of DBC \citep{DBC}, and evaluate the advantages of learning action representations on the task of Cheetah. The objective is to train the robot to be able to run in a straight line in a stable running stance, with the reward obtained being proportional to the forward speed. The state space is continuous and consists of classical RGB images with 84*84 pixels, and the action space is 6-dimensional and continuous. The reward function is defined as $r(v)=\max \left(0, \min \left(\frac{v}{10}, 1\right)\right)$, where $v$ is represented by velocity of the robot. In this task, we mainly validate the advantages of introducing action representation learning.

First, we use the constructed $V(a_t)$ to represent the action representation in latent space, where the dimensionality of action representation $e$ is set as $N_e=12$. Then, we learn the policy for the Cheetah task, which uses the state-of-the-art DBC method \citep{DBC} as the policy update method. Therefore, we compare the following two methods:
\begin{itemize}
	\item \textbf{DBC}: A task-relevant state representation method based on SAC policy learning algorithm \citep{DBC}.
	\item \textbf{PL-LS with DBC}: Implementation our proposed framework of policy
		learning in latent spaces with DBC, where state representation is learned by DBC and action representation is learned by $V_a$.
\end{itemize}

\begin{figure}[h]
	\centering
	\includegraphics[scale=.65]{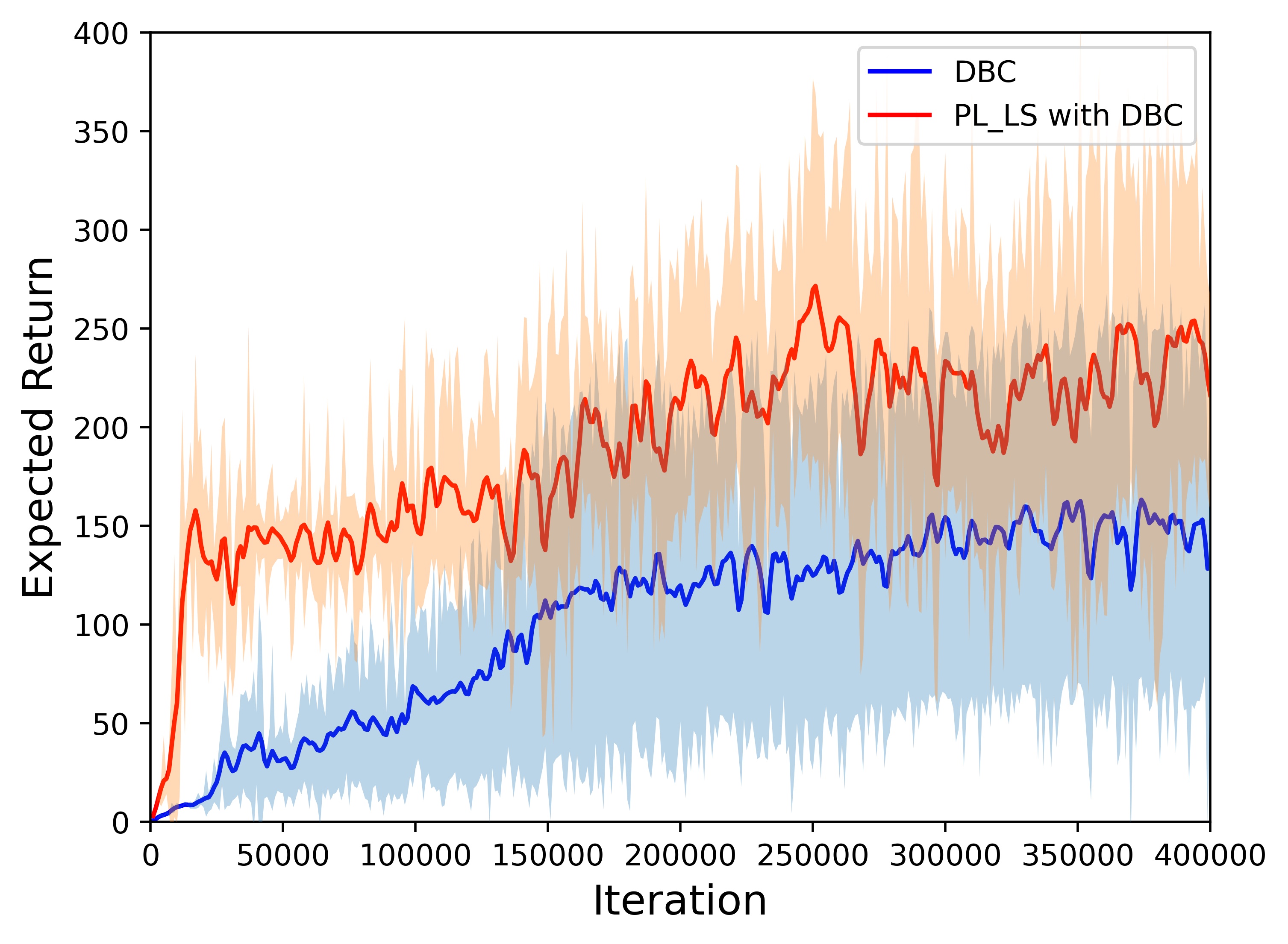}
	\caption{Average expected return over 5 trials for Cheetah. The horizontal coordinate indicates the number of model iterations and the vertical coordinate indicates the expected return. The shaded areas are the standard deviations.} \label{cheetahReturn}
\end{figure}

\begin{figure}[H]
	\centering
	\includegraphics[scale=.93]{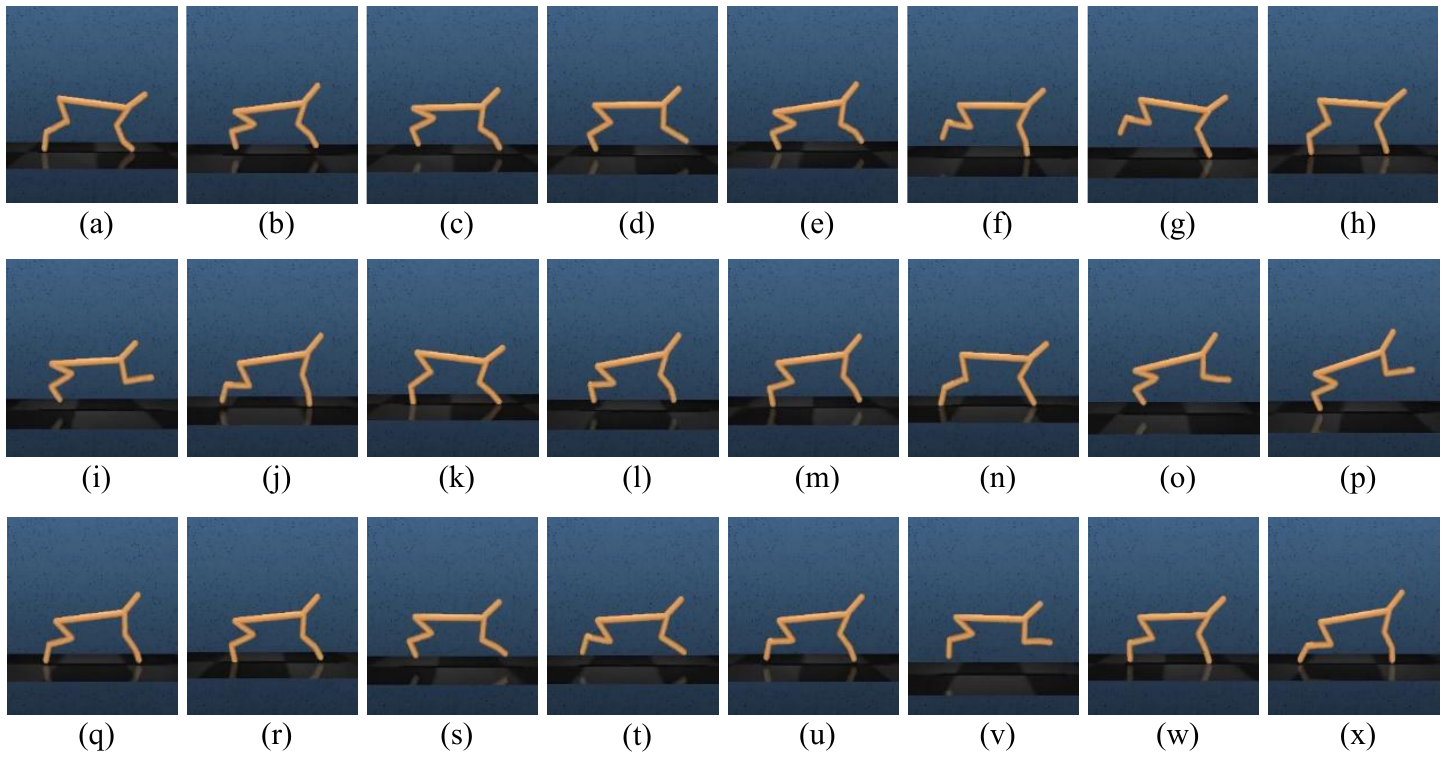}
	\caption{The demonstration of learned policy by PL-LS. The sampling interval is 40 frames.} \label{cheetah}
\end{figure}

We investigate the average return for Cheetah over 5 trials. The experimental results are plotted in \hyperref[cheetahReturn]{Fig.21}, which show that PL-LS with DBC outperforms the standard DBC, which obtain higher expected rewards and learn policy faster. Finally, we illustrate an example of policy obtained by PL-LS with DBC and the trajectory is plotted in \hyperref[cheetah]{Fig.22}, where the sampling interval is 40 frames. Through this illustration, we can see that the robot can run in a straight line with a stable stance. Therefore, the learning of the action representation proposed in PL-LS can improve the performance of policy learning.

\section{Conclusion and Future Work}
In this paper, we proposed an efficient policy learning framework in latent spaces (PL-LS), which included the state representation and action representation in the policy learning framework to improve the learning efficiency and generalization capability. More specifically, we built a simple and compact policy model to reduce the burden of policy learning by fusing the state representation space with the action representation space. Furthermore, the proposed PL-LS method was shown to improve the generalization performance of action selection by generalizing actions to other actions with similar representations through action representations. The effectiveness of the proposed idea was demonstrated through extensive experiments, which showed that the proposed PL-LS was promising.

To implement the proposed PL-LS method, the algorithm of PPO was employed, but it was not limited to the chosen one here; this model framework can be easily extended by using other policy learning methods, e.g. DDPG \citep{2015Continuous}, TRPO \citep{2015Trust}, A3C \citep{2016A3C} and SAC \citep{SAC} et al. Specifically, we implement a state-of-the-art method DBC in our proposed framework PL-LS and evaluate its performance in the Cheetah task, and the results show that it significantly improves the performance of the policy and learning efficiency. We trained the state and action representations in the offline manner for efficient policy learning, learning representation models in a different way, such as online and offline interlaced manner, may further improve the performance, which is an interesting direction.
Our approach was experimentally demonstrated to be feasible, but it was not yet proven from a theoretical perspective, which is something we need to explore further in our future work.

\section{Acknowledgments}

This work was supported by the National Natural Science Foundation of China [grant number 61976156]; Tianjin Science and Technology Commissioner project [grant number 20YDTPJC00560]; and Natural Science Foundation of Tianjin [grant number 18JCQNJC69800].


%%\bibliographystyle{elsarticle-harv}
\bibliography{mybibfile}




%%\bibitem[ ()]{}

%%\end{thebibliography}
\end{document}